\documentclass[referee,sn-apa,pdflatex,natbib]{sn-jnl}

\usepackage{graphicx}%
\usepackage{multirow}%
\usepackage{amsmath,amssymb,amsfonts}%
\usepackage{amsthm}%
\usepackage{mathrsfs}%
\usepackage[title]{appendix}%
\usepackage{xcolor}%
\usepackage{textcomp}%
\usepackage{manyfoot}%
\usepackage{booktabs}%
\usepackage{algorithm}%
\usepackage{algorithmicx}%
\usepackage{algpseudocode}%
\usepackage{listings}%
\usepackage{csquotes}
\usepackage{lmodern}
\usepackage{anyfontsize}
\usepackage{longtable}
\usepackage{supertabular}
\usepackage{caption}

\begin{document}

\title[Article Title]{Bird Eye-View to Street-View: A Survey}


\author[1]{\fnm{Khawlah} \sur{Bajbaa}}\email{g202115030@kfupm.edu.sa}

\author[3]{\fnm{Muhammad} \sur{Usman}}\email{muhammad.usman8@ontariotechu.ca}
\equalcont{These authors contributed equally to this work.}

\author[1,2]{\fnm{Saeed} \sur{Anwar}}\email{saeed.anwar@kfupm.edu.sa}
\equalcont{These authors contributed equally to this work.}

\author[4]{\fnm{Ibrahim } \sur{Radwan}}\email{ibrahim.radwan@canberra.edu.au}
\author[5]{\fnm{Abdul Bais}\sur{}}\email{abdul.bais@uregina.ca}

\affil[1]{\orgdiv{Department of Information and Computer Science}, \orgname{King Fahd University of Petroleum and Minerals}, \city{Dhahran}, \postcode{31261}, \country{Saudi Arabia}}

\affil[2]{\orgdiv{SDAIA-KFUPM Joint Research Center of Artificial Intelligence}, \city{Dhahran}, \postcode{31261}, \country{KSA}}

\affil[3]{\orgdiv{Faculty of Science}, \orgname{Ontario Tech University}, \city{Oshawa}, \postcode{L1G 0C5}, \country{Canada}}

\affil[4]{\orgdiv{Faculty of Science and Technology}, \orgname{University of Canberra}, \city{ACT}, \postcode{2601}, \country{Australia}}

\affil[5]{\orgdiv{Electronic Systems Engineering}, \orgname{University of Regina}, \orgaddress{\city{Saskatchewan}, \postcode{3737}, \country{Canada}}}


\abstract{In recent years, street view imagery has grown to become one of the most important sources of geospatial data collection and urban analytics, which facilitates generating meaningful insights and assisting in decision-making. Synthesizing a street-view image from its corresponding satellite image is a challenging task due to the significant differences in appearance and viewpoint between the two domains. In this study, we screened 20 recent research papers to provide a thorough review of the state-of-the-art of how street-view images are synthesized from their corresponding satellite counterparts. The main findings are: (i) novel deep learning techniques are required for synthesizing more realistic and accurate street-view images; (ii) more datasets need to be collected for public usage; and (iii) more specific evaluation metrics need to be investigated for evaluating the generated images appropriately. We conclude that, due to applying outdated deep learning techniques, the recent literature failed to generate detailed and diverse street-view images.}




\maketitle

\section{Introduction}\label{sec1}

Generating street-view images from satellite-view is one of the rising topics in the cross-view image translation domain. It aims to generate geometrically consistent street views corresponding to the center-located area on the provided satellite view. As satellite images are becoming more widely available (\textit{e.g.}, Google Earth), it is becoming easier to cover almost every spot of the globe. However, such a capability does not exist for street-view images~\cite{lu2020}, as the majority of street views are captured around known landmarks and major roads~\cite{workman2015wide},~\cite{crandall2009mapping}. It helps enhance the content in areas that are expensive or difficult to reach by people or vehicles. Also, street-view images enable more accurate geo-localization than satellite-view images.

Street-view synthesis is helpful for various downstream tasks, such as image geo-localization, which is considered a fundamental problem in many applications such as robotics, augmented reality, and autonomous driving~\cite{toker2021}. The retrieval component is limited to identifying and locating similar objects, so adopting satellite to street-view synthesis plays a vital role in this task, where the generator can learn more valuable features that help in precise location retrieval. Moreover, street-view images have many additional useful details not in satellite images, such as building facades~\cite{toker2021}. Furthermore, cross-view matching between satellite and street views helps in predicting street-view segmentation maps based on satellite views of the same location for geo-orientation and geo-location estimation tasks by extracting features from a satellite image and mapping them into street-view, conveying semantics from street-view to satellite view~\cite{zhai2017}. Moreover, the representation derived from the conditional Generative Adversarial Networks (GANs) for generating street views from satellite images aids in distinguishing between rural and urban land covers~\cite{deng2018}.

In recent years, synthesizing street-view from satellite images has been investigated by Generative Adversarial Network-based methods~\cite{shi2022},~\cite{wu2022cross},~\cite{deng2018},~\cite{ren2021cascaded},~\cite{toker2021}, or by combining U-Net~\cite{ronneberger2015} and BicycleGAN, such as~\cite{lu2020}. Differently,~\cite{regmi2018} and~\cite{wu2022cross} explored the conditional GAN that learns the generation of both images, either street-view or satellite-view, and their corresponding semantic maps simultaneously. In addition,~\cite{Tang_2020_CVPR} and~\cite{tang2019multi} investigated the semantic-guided cGAN where the semantic map is concatenated with the input image. As a result, the semantic map serves as a guide to direct the model in generating images in another domain. On the other hand, street-view to satellite synthesis, has been explored by employing a conditional GAN \cite{regmi2019bridging} for street-view geo-localization task for predicting street-view location. In addition, some other works investigated the street-view geo-localization task by matching it with satellite images using Convolutional Neural Networks (CNN) and feature matching-based methods~\cite{shi2020looking},~\cite{workman2015wide}, Siamese Network~\cite{Lin_2015_CVPR},~\cite{hu2018},~\cite{Liu_2019_CVPR}, and Siamese and Triplet Network~\cite{vo2016localizing}.  Recently, more advanced methods are being used for satellite to street-view synthesis domain, such as transformer-based \cite{swerdlow2024street} and diffusion-based methods \cite{li2024sat2scene}, \cite{li2024crossviewdiff}.

\begin{figure}[tbp]
    \centering
    \includegraphics[width=0.99\linewidth]{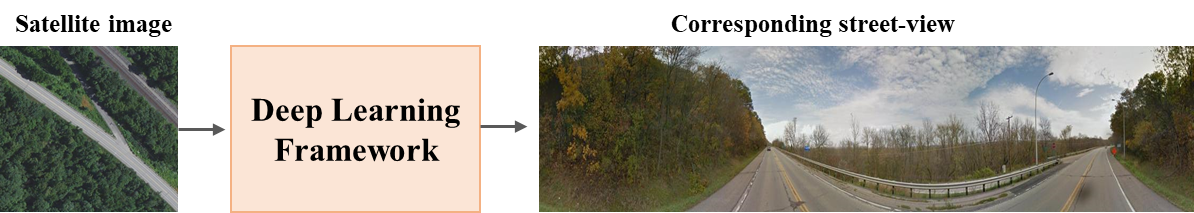}
    \caption{General overview of satellite to street-view generation. }
    \label{fig:general-overview}
\end{figure}

Figure~\ref{fig:general-overview} illustrates a general overview of the satellite to street-view generation. As a result of the challenging nature of this problem, it leaves an opportunity for further improvements. Despite the progress of satellite-to-street-view synthesis, it has limitations, as a satellite image only shows the top surface of an object. Therefore, it led to a hallucinatory representation of the facades of the objects. In addition, indistinguishable small objects such as cars, tree shadows, sidewalks, and fences are still being ignored in street-view synthesis.

\begin{figure*}[tbp]
    \centering
    \includegraphics[width=0.9\textwidth]{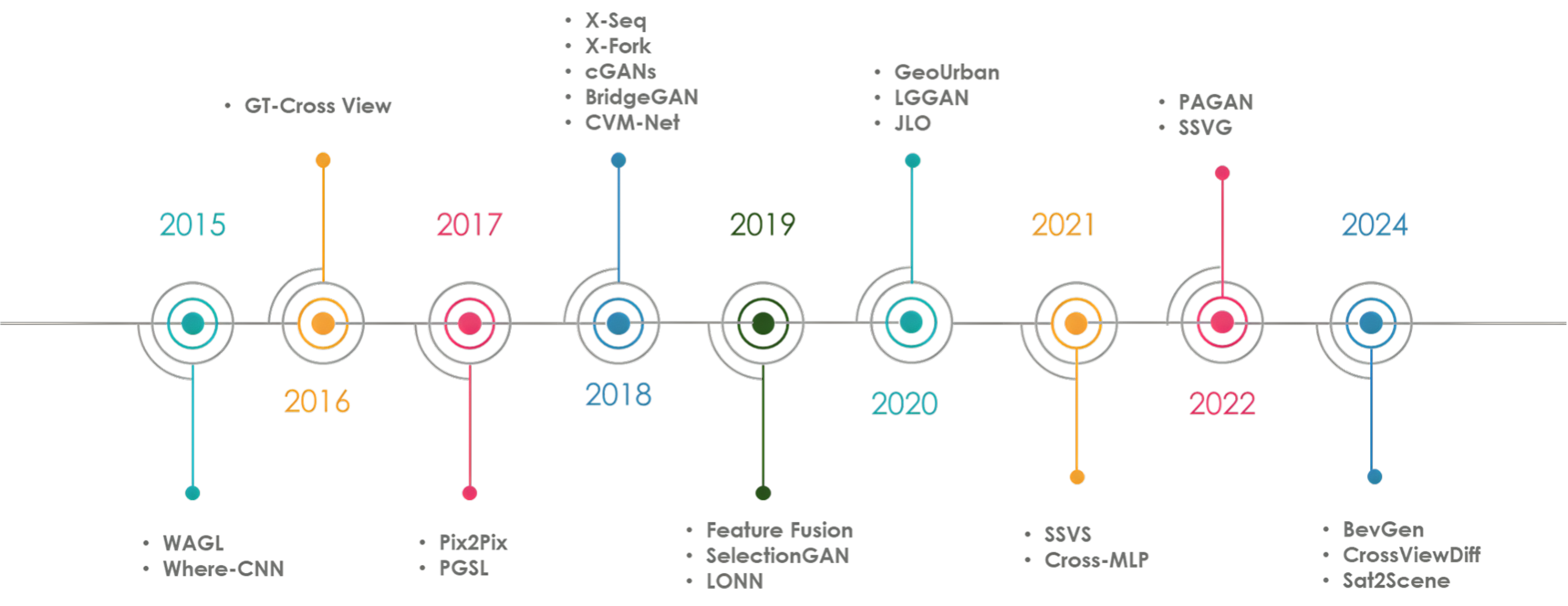}
    \caption{Chronological display of the papers included in the survey. The chart illustrates the increasing trend towards deep models in the research evolving in the area of birds eye view to street view.}
    \label{fig:timeline}
\end{figure*}

\begin{figure*}[tbp]
    \centering
    \includegraphics[width=0.9\textwidth]{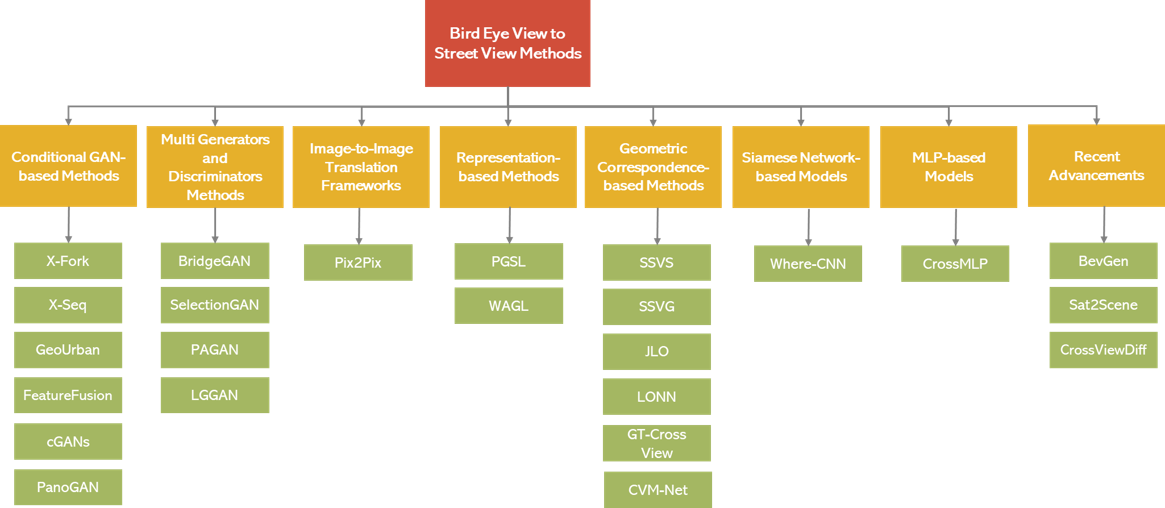}
    \caption{Taxonomy diagram of bird eye view to street-view existing methods, showing the classification and corresponding methods within each category. }
    \label{fig:taxonomy_diagram}
\end{figure*}

\section{Bird Eye View to Street View Methods}\label{sec2}

\textcolor{black}{This section reviews various methods in the bird-eye view to street-view domain. We categorize each method based on the applied techniques. Figure \ref{fig:taxonomy_diagram} provides an overview for these methods and their corresponding categories, and Table \ref{tab:summary-LR} presents a summary of the reviewed methods. }

\subsection{Conditional GAN-based Models}

These methods share the commonality of utilizing conditional Generative Adversarial Networks (GANs) for image synthesis tasks. They are designed to generate realistic images from different viewpoints or modalities, such as from bird's eye to street view or vice versa. The focus is leveraging conditional information to produce high-quality images exhibiting desired characteristics.

\begin{figure*}[tbp]
\centering
\begin{tabular}{cc}
\includegraphics[width=0.9\columnwidth]{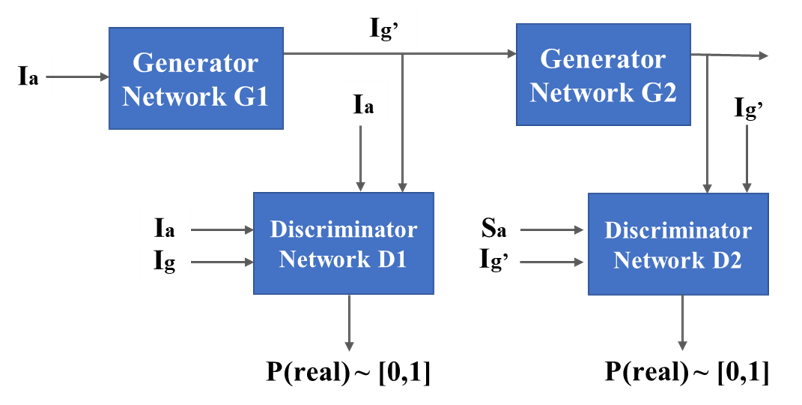}& 
\includegraphics[width=0.9\columnwidth]{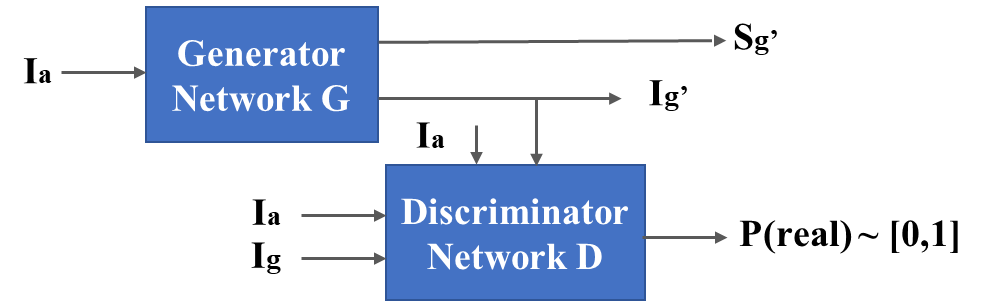}\\
(a) X-Seq~\cite{regmi2018}  &
(b) X-Fork~\cite{regmi2018}\\

\multicolumn{2}{c}{\includegraphics[width=15cm, height=5cm]{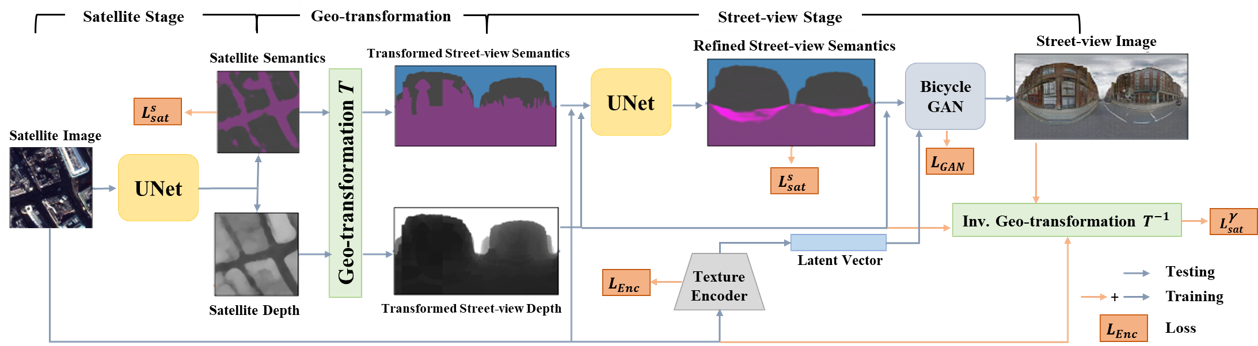}} \\
\multicolumn{2}{c}{(c) GeoUrban~\cite{lu2020}}\\

\includegraphics[width=0.9\columnwidth]{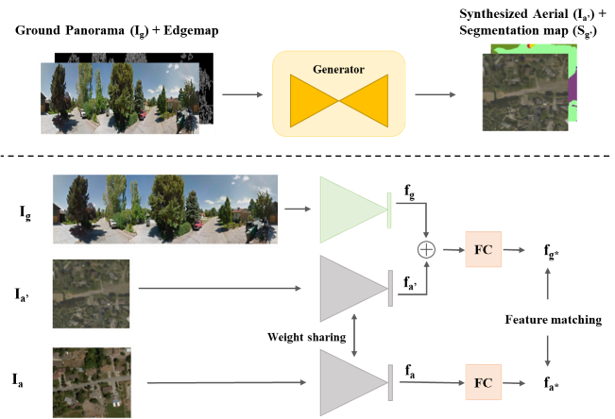} &
\includegraphics[width=10cm, height=4cm]{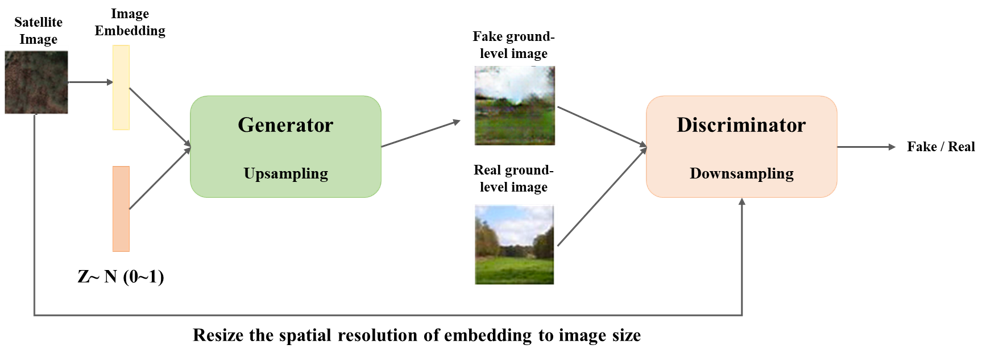} \\
(d) FeatureFusion~\cite{regmi2019bridging}
&


(e) cGANs~\cite{deng2018}\\
\multicolumn{2}{c}{\includegraphics[width=15cm, height=5cm]{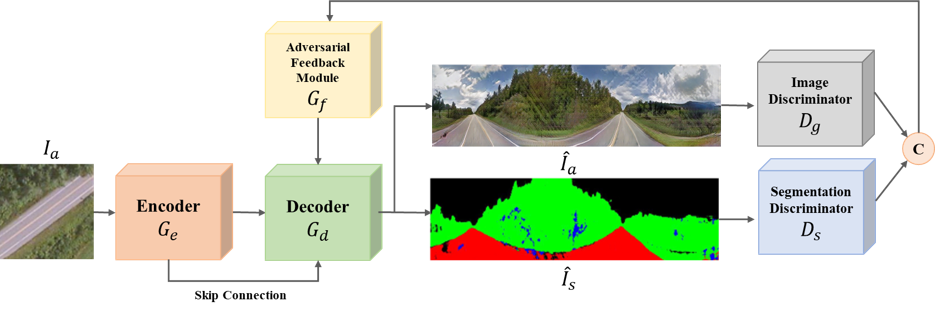}}\\
\multicolumn{2}{c}{(f) PanoGAN~\cite{wu2022_panogan}}
\end{tabular}
\caption{Conditional GAN-based models.}
\label{fig:cGANmodels}
\end{figure*}

\vspace{1mm}
\noindent\textbf{X-Fork~\cite{regmi2018}:} Regmi and Borji introduced two architectures based on conditional GANs called Cross-view Fork (X-Fork) and Sequential-view (Seq-X) for cross-image synthesis from street view to aerial view and vice versa. The X-Fork architecture used the baseline discriminator, where the generator network is forked, to produce segmentation maps and images, as shown in Figure~\ref{fig:cGANmodels}(b). The Seq-X architecture uses two conditional GAN networks:  the first network generates baseline cross-view images, which are fed to the second network as conditioning input to create the segmentation map in the same view, as illustrated in Figure~\ref{fig:cGANmodels}(a). The results illustrated that X-Fork achieved higher accuracy with Top-1 and Top-5 for 64$\times$64 resolution. On the other hand, X-Seq performed better with 256$\times$256 resolution. This study employs two datasets: the first is the Dayton~\cite{vo2016localizing}, and the second is the CVUSA.

\vspace{1mm}
\noindent\textbf{GeoUrban* \cite{lu2020}:} It exploits geometry to synthesize street-view images from satellite images. This study shows a 5km by 5km region of interest in the center of London City. The ground truth of depth and semantic satellite images is generated from stereo matching and supervised classification with post-correction. The architecture consists of three stages, as shown in Figure~\ref{fig:cGANmodels}(c). The first is the satellite stage, which uses U-Net~\cite{ronneberger2015} to calculate a given satellite image's depth and semantic images. The second stage is the geo-transformation stage, which converts the semantic and depth images of satellite images into their corresponding semantic and depth panorama images. The third stage, the street-view stage, is another U-Net~\cite{ronneberger2015} that refines the semantic panorama. Eventually, the BicycleGAN~\cite{zhu2017toward} is applied to generate realistic images from the semantic panorama. The novelty of the proposed method lies in offering a new transformation module for generating street-view images constrained by the geometric information of satellite images. Also, a novel weighted mask is utilized to mitigate the differences between the geometric panorama and satellite-depth images. The results demonstrated that the GeoUrban~\cite{lu2020} outperformed other methods like Pix2Pix~\cite{isola2017} and Regmi~\textit{et~al.}~\cite{regmi2018}'s model with an accuracy of 72.9\%.

\vspace{1mm}
\noindent\textbf{FeatureFusion~\cite{regmi2019bridging}:} It takes advantage of conditional GANs to synthesize aerial images from ground-level panoramas. The X-Seq \cite{regmi2018} is adopted with some modifications, such as reshaping the features into $2\times2$ and then performing different up-convolution operations to generate $512\times512$ aerial image resolution. The ground-level image and its edge-map are inputs to the X-Fork generator~\cite{regmi2018}, as shown in Figure~\ref{fig:cGANmodels}(d),  which generates a synthesized aerial image and its corresponding edge-map. After that, FeatureFusion learns the joint feature representation of the ground-level image and the synthesized aerial image, and the features of the three images are integrated, followed by a fully connected layout to get a robust representation that matches the aerial feature representation. The novelty of the method lies in combining the edge map and semantic segmentation with input images to enhance the synthesis of cross-view. The results showed that the proposed feature fusion-based outperformed other approaches, including~\cite{workman2015wide,zhai2017,vo2016localizing,hu2018} with 48.75\% top-1 recall, 81.27\% top-10 recall, and 95.98\% top-1 accuracy. Moreover, FeatureFusion~\cite{regmi2019bridging} experiments on the CVUSA and a new dataset the authors collected, the Orlando-Pittsburgh (OP) dataset.

\vspace{1mm}
\noindent\textbf{cGANs~\cite{deng2018}:} use conditional GANs for generating ground-level images from satellite images and investigated cGANs feature maps for the land-cover classification. The satellite photos are collected via the Google Map Statistics API, while the ground-level images are taken from the Geograph API with known locations. The proposed framework addresses the problem of expressing an aerial picture's information on how things appear on a ground-level image so that it can be used as input to the cGANs to generate ground-level photos. However, feeding 2D satellite image vectors to cGAN is ineffective due to the lack of structural similarity between satellite and ground-level images. Therefore, three types of image patch embedding are utilized to capture structural similarity: grayscale, HSV, and CNN. Figure~\ref{fig:cGANmodels}(e) shows the proposed framework where the 1D image embedding vector is joined with a noise vector for the generator's input. The feature maps generated by cGANs achieved the highest accuracy, with a value of 73.14\%.

\vspace{1mm}
\noindent\textbf{PanoGAN~\cite{wu2022cross}:}Wu \textit{et al.} \cite{wu2022cross} proposed an innovative adversarial feedback GAN-based framework (PanoGAN) designed satellite to street-view synthesis task. It consists of two main components: a dual-branch discriminator and adversarial feedback module. During the training phase, the segmentation maps have been incorporated into the proposed framework to guide the street-view generation process and enhance the geometric consistency. PanoGAN has been designed in a way to improve the generator-discriminator cooperation during the generation process. First, the satellite image is input into the generator to generate the panorama street-view image and its corresponding segmentation map. Then, the discriminator’s feature responses are encoded through the adversarial model and fed it back into the generator, to produce the refined version of panorama street-view and segmentation map image pair. This iterative process allows continuous improvements by leveraging discrimination feedback as a powerful source for enhancing the panorama street-view images and their associated segmentation maps. The novelty of this work lies in proposing the dual-branch discrimination strategy to ensure semantic consistency and high-quality images.  In addition, the adversarial feedback module is introduced to improve the generation process. The results illustrate that, PanoGAN reached to the highest scores on CVUSA dataset with SSIM, PSNR, and KL values of 0.4437, 20.9467, and 4.20±1.19, respectively. On Orlando-Pittsburgh (OP) dataset it reached to the highest scores on OP dataset in terms of SSIM, KL and SD with values of 0.4428 , 7.89±1.12, and 18.0543, respectively. Figure \ref{fig:cGANmodels} (f) shows an overview of PanoGAN architecture.

\subsection{Multi Generator and Discriminator Models}
As the name indicates, the multiple generator and discriminator models hierarchically employ many generators and discriminators.

\begin{figure*}[tbp]
\centering
\begin{tabular}{cc}
\includegraphics[width=0.45\linewidth]{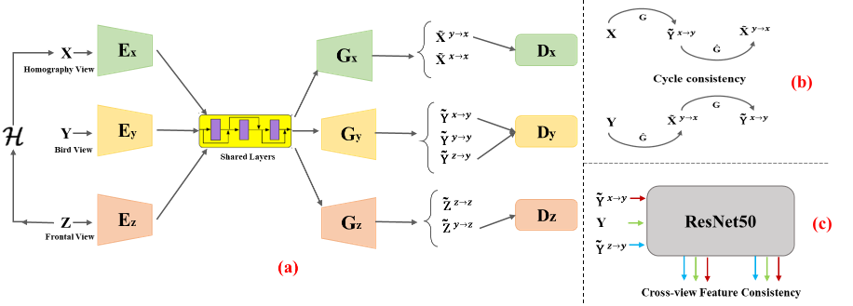}&
\includegraphics[width=0.45\linewidth]{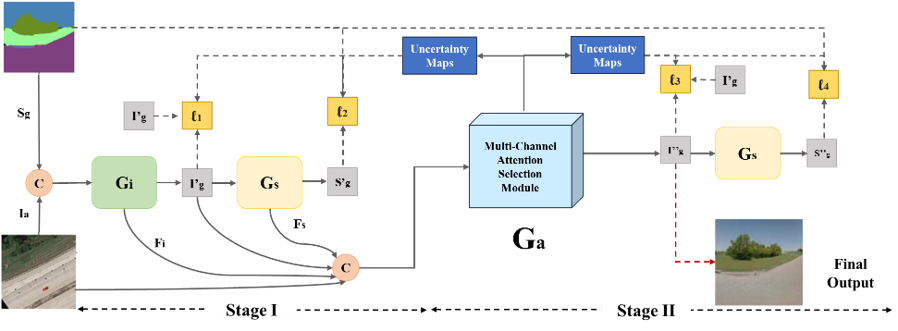}\\
(a) BridgeGAN~\cite{zhu2018generative}&
(b) SelectionGAN~\cite{tang2019multi}\\

\multicolumn{2}{c}{\includegraphics[width=15cm, height=5cm]{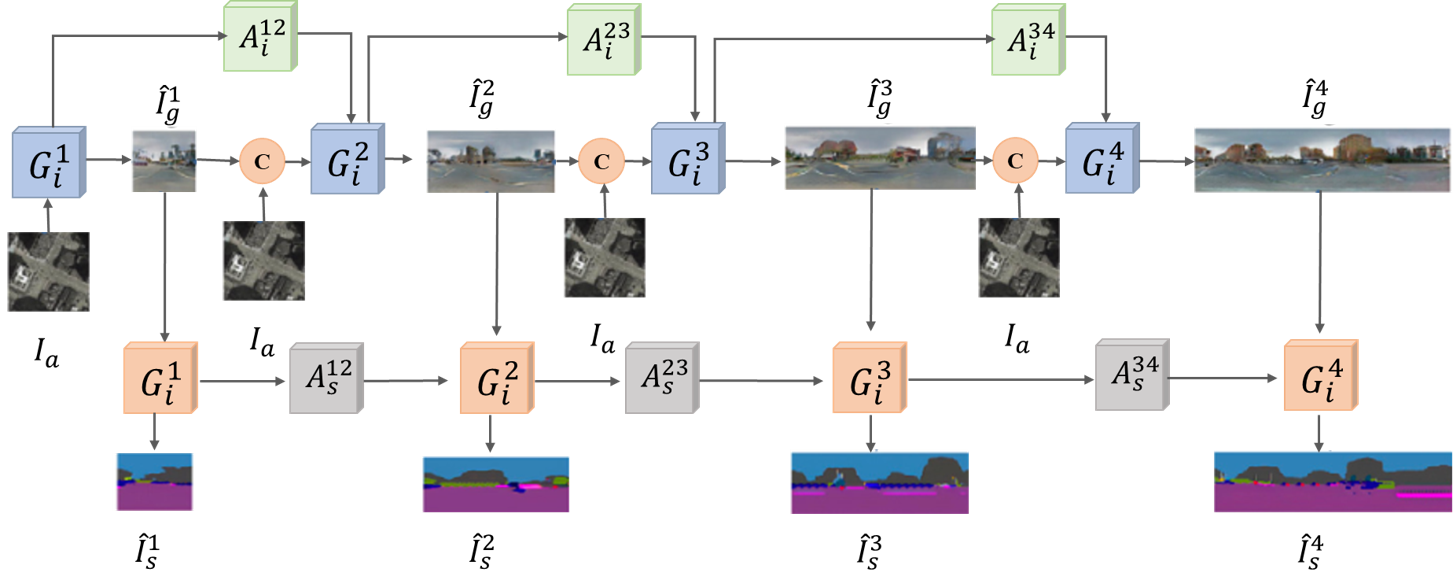}}\\
\multicolumn{2}{c}{(c) PAGAN~\cite{wu2022cross}} \\

\multicolumn{2}{c}{\includegraphics[width=15cm, height=5cm]{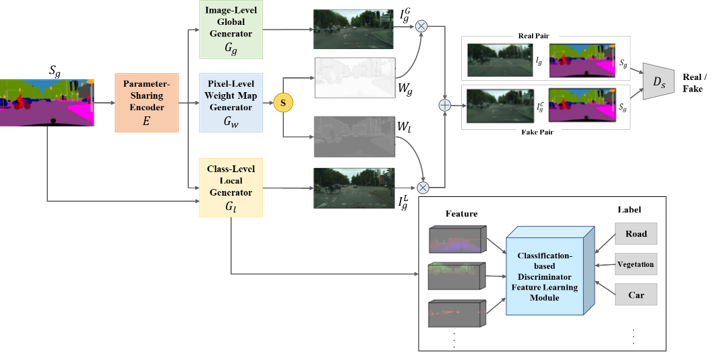}}\\

\multicolumn{2}{c}{(d) LGGAN~\cite{Tang_2020_CVPR}} \\

\end{tabular}
\caption{Multi generator and discriminator models.}
\label{fig:multiGAN}
\end{figure*}

\vspace{1mm}
\noindent\textbf{BridgeGAN~\cite{zhu2018generative}:} A multi-GAN model for generating bird-view images, BridgeGAN, introduces an intermediate view called the homography view to narrow the gap between the bird view and the frontal view. BridgeGAN consists of three GAN networks, as shown in Figure~\ref{fig:multiGAN}(a), each representing a specific view domain, namely homography-view, bird-view, and frontal-view. Each view domain has three components: a generator, a discriminator, and an encoder. In addition, shared layers are used to learn integrated intermediate representations since all three domains have the same semantic meaning. Cycle consistency loss consists of backward and forward cycles, while the introduced RestNet50 loss network imposes cross-view feature consistency. The novelty of BridgeGAN lies in addressing bird-view image generation from frontal-view images. The BridgeGAN achieves the best performance with values of 0.596 for SSIM, 5.01 for PSNR, and 0.242 for LPIPS compared with other baselines: Pix2Pix~\cite{isola2017}, CycleGAN~\cite{zhu2017unpaired}, DiscoGAN~\cite{kim2017learning}, and CoGAN~\cite{liu2016co}. The dataset is obtained by collecting images from the Grand Theft Auto V video game using Palazzi \textit{et al.}'s~\cite{palazzi2017learning} framework.

\vspace{1mm}
\noindent\textbf{SelectionGAN~\cite{tang2019multi}:} A multi-channel attention selection GAN comprises two stages for synthesizing images conditioned on the reference image and semantic map, as illustrated in Figure~\ref{fig:multiGAN}(b). The first stage focuses on the semantic features of the views, takes the target semantic map and the condition image (i.e., the ground-level image), and produces initial coarse results. The second stage targets the appearance details by refining the initial results via the multi-channel attention selection mechanism. The novelty lies in using coarse-to-fine inference to investigate cascaded semantic guidance. Moreover, a multi-attention module is offered to choose the intermediate generation attentively, which helps improve output quality. SelectionGAN~\cite{tang2019multi} model is evaluated on three datasets: CVUSA~\cite{workman2015wide} for aerial to ground-level images, Dayton~\cite{vo2016localizing} for ground-level to aerial images, and Ego2Top \cite{ardeshir2016ego2top} (consists of pairs of two images of the same scene from two different views). The SelectionGAN achieved the best performance compared to Pix2Pix~\cite{isola2017}, X-Fork~\cite{regmi2018}, X-Seq~\cite{regmi2018}, and PGSL~\cite{zhai2017}. In some cases, on the CVUSA dataset, the X-Seq~\cite{regmi2018} occasionally performs slightly higher; however, SelectionGAN generates more realistic images.

\vspace{1mm}
\noindent\textbf{PAGAN~\cite{wu2022cross}:} Progressive Attention GAN has two primary components: a cross-stage attention module and a multi-stage progressive generation framework. Figure~\ref{fig:multiGAN}(c) shows an overview of the PAGAN, in which cross-stage attention is applied to generate a panorama through several coherent stages. PAGAN~\cite{wu2022cross} comprises multiple segmentation and image generators for generating the street-view panorama and its segmentation map. In the first stage, the aerial image is fed to the image generator and segmentation generator, which produce the panorama and its corresponding segmentation map. After that, the generated image is refined by applying the cross-stage attention module, which forwards the semantic information stage by stage. The novelty of PAGAN lies in proposing a progressive generative framework for the high-resolution generation of street-view panoramas. Also, introducing a new cross-stage attention mechanism to bridge the subsequent generational stages allows the quality of street-view panoramas to improve continuously. The PAGAN outperformed other recent works: Pix2Pix~\cite{isola2017}, X-Fork~\cite{regmi2018}, X-Seq~\cite{regmi2018}, SelectionGAN~\cite{tang2019multi}, GAS2G~\cite{lu2020}, especially in terms of KL and FID scores, where the PAGAN reduced these scores significantly. In this study, two datasets are used, namely, the CVUSA and Orlando-Pittsburgh (OP) datasets.

\vspace{1mm}
\noindent\textbf{LGGAN~\cite{Tang_2020_CVPR}:} A Local class-specific and Global image-level GANs (LGGAN) model addresses semantic-guided-based scene generation tasks. LGGAN generation module is composed of three parts: a class-specific local generator for enriching the local scene details, an image-level global generator for capturing target images' overall structure and layout, and a weight-map generator for combining the resulted images from the global and local generators with the learned weight maps, as shown in Figure~\ref{fig:multiGAN}(d). The discriminator tries to differentiate generated images from semantic and image space. In addition, a new classification module is proposed to learn better discriminative class-specific feature representation. The novelty of LGGAN is that it considers both local and global contexts. The  LGGAN performs better on most evaluation metrics compared with other recent works, such as pix2pix~\cite{isola2017}, SelectionGAN~\cite{tang2019multi}, X-Seq~\cite{regmi2018}, and X-Fork~\cite{regmi2018}. While SelectionGAN achieved slightly better performance on pixel-level metrics, i.e., PSNR, SD, and SSIM, than LGGAN. However, LGGAN generates more visually detailed objects like trees, buildings, and cars than SelectionGAN. The image translation part has been evaluated using the Dayton and CVUSA  datasets.

\subsection{Image-to-Image Translation Frameworks}
Image-to-image translation frameworks operate on the premise of learning mappings between input and output images through supervised learning with paired data.

\vspace{1mm}
\noindent\textbf{Pix2Pix~\cite{isola2017}:} A general framework that investigates using the conditional GAN for image-to-image translation. Though Pix2Pix~\cite{isola2017} is not specific to bird-view to street-view, it has been extensively cited in this domain. The Pix2Pix~\cite{isola2017} method is evaluated on many tasks and datasets: i) semantic labels to photos trained on Cityscapes~\cite{cordts2016cityscapes}, ii) map to aerial employs Google Map images, iii) edges to photo for~\cite{zhu2016generative} and~\cite{yu2014fine}, iv) sketch to photo for human sketches utilizing~\cite{eitz2012humans}, and v) day to night using~\cite{laffont2014transient} data. The architecture of the generator and discriminator has been adapted from~\cite{radford2015unsupervised}, where both are composed of Conv-BN-ReLU~\cite{ioffe2015batch}. The skip connections are added to the generator, following the general shape of the U-Net~\cite{ronneberger2015}, to allow low-level information from initial layers. Moreover, a novel discriminator, termed PatchGAN, penalizes structure at the image patch scale. The results showed that Pix2Pix~\cite{isola2017} achieved the second-best result on the colorization task compared with Colorful Colorization~\cite{zhang2016colorful}, and a variant of their approach employed $\ell_2$ regression~\cite{zhang2016colorful}. Additionally, the Pix2Pix~\cite{isola2017} model's aerial image generated from the map deceived the participants far more than the baseline. On the other hand, the generated map from an aerial image misled the participants by 6.1\%, which is not substantially exceeding the baseline.

\subsection{Representation-based Models}
These models prioritize extracting meaningful features from aerial photos and ground-level images, also called representations that capture essential semantic information. These representations serve as a compact and informative encoding of the input data, which can then be utilized for various downstream tasks. Representation learning involves training neural networks to learn features from raw image data automatically. The features are learned to capture relevant characteristics of the scenes depicted in the images, such as spatial layouts, object categories, textures, and other semantic information.

\begin{figure}[tbp]
    \centering
    \includegraphics[width=0.99\linewidth]{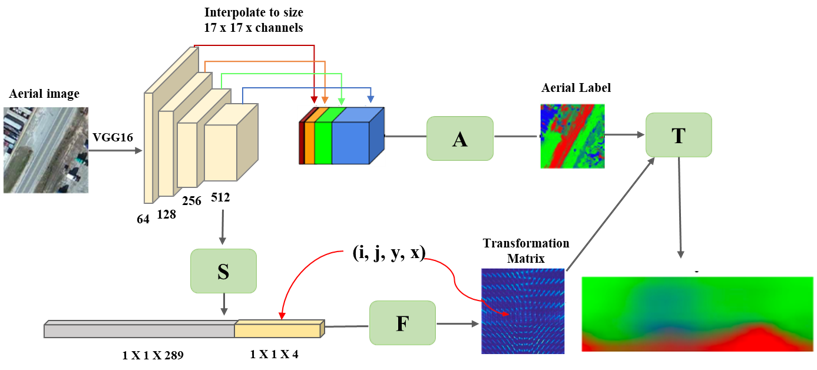}
    \caption{PGSL: Semantic Feature Learning and Representation-based Models~\cite{zhai2017}. }
    \label{fig:PGSL}
\end{figure}

\vspace{1mm}
\noindent\textbf{PGSL*~\cite{zhai2017}:} Zhai~\textit{et~al.} presented a novel training technique for extracting meaningful features from aerial photos. The aim is to utilize the associated aerial image to predict the ground-level semantic layout. Four networks handle the features, as shown in Figure~\ref{fig:PGSL}. The CNN network~\enquote{A} extracts features from the aerial image via VGG16, and then it is fed to PixelNet~\cite{bansal2016pixelnet} to convert it to pixel-level. The CNN network, \enquote{S}, employs the extracted features from aerial images to control the transformation. The network~\enquote{F} estimates the change between viewpoints. Ground-level semantic labeling is created by applying the transformation, \enquote{T}, to the aerial semantic features. The novelty of this study lies in a new CNN design that links the appearance of an aerial image to a ground-level semantic layout located in the same location.
Further, the PGSL~\cite{zhai2017} method can be extended to ground-level image orientation estimation, synthesis, and localization. Three different initializations have been experimented with: the proposed method initialization, Xavier~\cite{glorot2010understanding} random initialization, and the VGG16 pre-trained on ImageNet initialization. The results demonstrated a superior performance of the proposed method compared to other initialization techniques on the CVUSA benchmark dataset.

\vspace{1mm}
\noindent\textbf{WAGL*~\cite{workman2015wide}:} Workman~\textit{et~al.} address the wide-area geo-localization (WAGL) task employing pre-existing CNNs to extract the features from ground-level images and then learn to predict those features from the aerial photos with the exact location learning the semantic feature representation jointly. WAGL consists of two training models: a single-scale model and a multi-scale model. The first model primarily aims to extract aerial features corresponding to ground-level images. The second model matches ground-level images to aerial photos at multiple spatial scales. The novelty lies in the in-depth analysis of existing CNN architectures for cross-view localization tasks with cross-view training for learning semantic features jointly from multiple image sources. The results showed that the features learned through multi-scale training outperformed single-scale and other methods~\cite{workman2015location,zhou2014} on Charleston and San Francisco benchmark datasets. 

\subsection{Geometric Correspondence Models}
The models in this group explicitly handle geometric correspondences between different views or modalities. They address the alignment of images based on known or learned geometric transformations, which is crucial for accurate information transfer and feature correspondence across domains. These methods ensure that the spatial relationships between images are preserved during synthesis or matching tasks.

\vspace{1mm}
\noindent\textbf{SSVS*~\cite{toker2021}:} It is a multi-task image synthesizing and retrieval architecture. SSVS stands for satellite-to-street view synthesis for geo-localization. The network learns the latent representations that are helpful for retrieval if they are employed for generating satellite and street-view images. In the proposed network, the generator is designed as U-Net~\cite{ronneberger2015}, which consists of residual blocks, while the discriminator is constructed as a PatchGAN classifier~\cite{isola2017}. The retrieval network consists of two main components, the encoder block and the spatial attention module, which convert an image's local features into global representation, as illustrated in Figure~\ref{fig:Geometrics}(a). The results showed that the proposed method outperformed other models such as Pix2Pix~\cite{isola2017}, X-Fork~\cite{regmi2018}, X-Seq~\cite{regmi2018}, and PGSL~\cite{zhai2017} on CVUSA dataset.

\begin{figure*}[tbp]
\centering
\begin{tabular}{cc}
\includegraphics[width=\columnwidth]{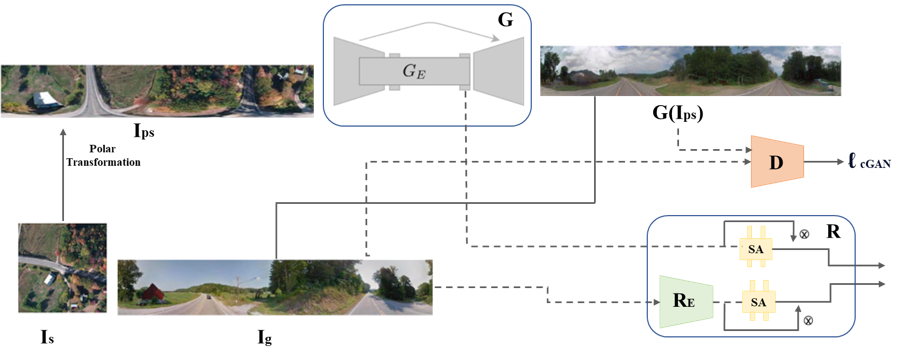}& 
\includegraphics[width=\columnwidth]{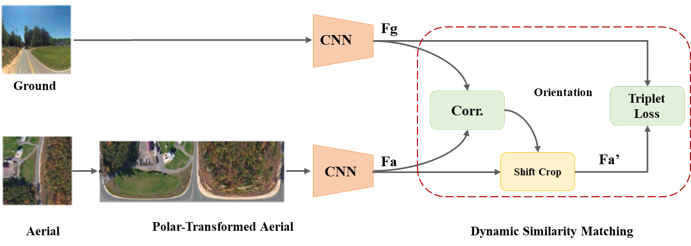}\\
(a) &
(b)\\

\includegraphics[width=0.9\columnwidth]{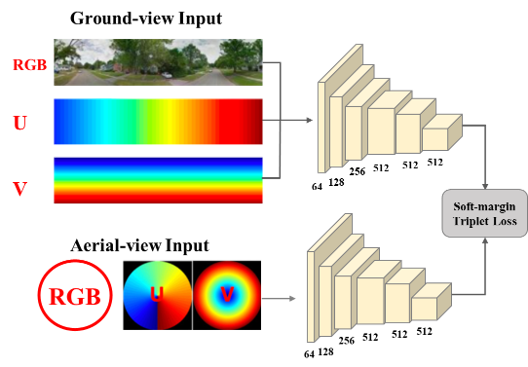}& 
\includegraphics[width=0.9\columnwidth]{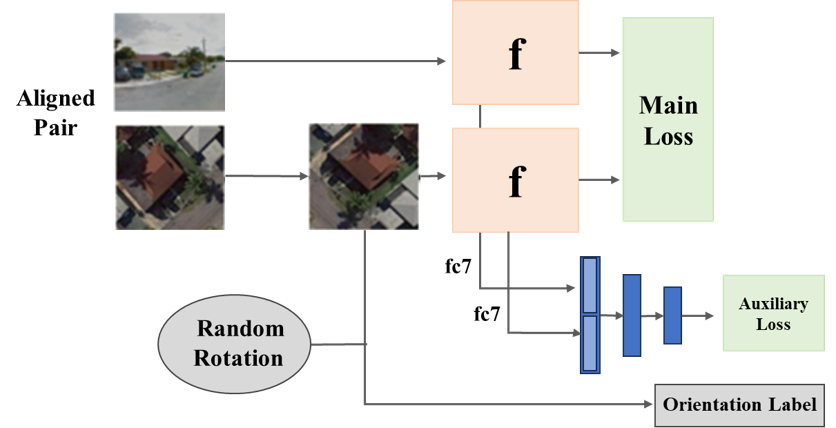}\\
(c) &
(d) \vspace{1mm}\\

\multicolumn{2}{c}{\includegraphics[width=0.7\textwidth]{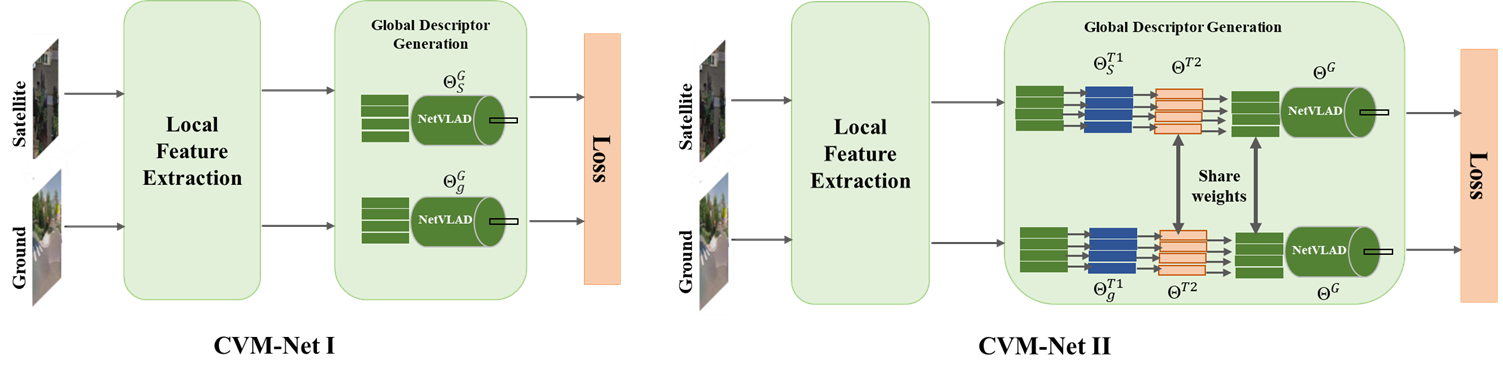}} 
\\
\multicolumn{2}{c}{(e)} \\
\end{tabular}
\caption{Geometric correspondence models. (a) SSVS~\cite{toker2021}, (b) 
JLO~\cite{shi2020looking}, (c) LONN~\cite{Liu_2019_CVPR}, (d)
GT-CrossView~\cite{vo2016localizing}, and (e) CVM-Net~\cite{hu2018}.}
\label{fig:Geometrics}
\end{figure*}

\vspace{1mm}
\noindent\textbf{SSVG*~\cite{shi2022}:} SSVG (Satellite and Street View Geometry) generates street-view images from satellite views based on modeling the geometric correspondences. Shi~\textit{et~al.}~\cite{shi2022} explicitly states the geometric correspondences between the satellite and street-view images, easing the information transfer between the mentioned domains. The proposed model applied the Satellite to Street-View Image Projection (S2SP) module to determine a geometric correspondence between street-view and satellite images. First, given the satellite image, the height probability is estimated through Pix2Pix~\cite{isola2017}, which is used for height estimation. Next, a satellite-view multiplane image (MPI) is constructed, given the satellite-view image and the estimated height probability distribution. Stretching and unrolling occur on each concrete cylinder to obtain a consistent street view. From street-view MPI, the final street-view image is generated in back-to-front order. Results are presented on two benchmark datasets: CVUSA~\cite{workman2015wide} and CVACT~\cite{Liu_2019_CVPR} and illustrated that the SSVG~\cite{shi2022} outperformed Pix2Pix~\cite{isola2017} and X-Fork~\cite{regmi2018} on both the datasets.

\vspace{1mm}
\noindent\textbf{JLO~\cite{shi2020looking}:} A recently proposed joint localization and orientation estimation method by Shi~\textit{et~al.}~\cite{shi2020looking} is based on Dynamic Similarity Matching (DSM) to measure the feature similarity of the image pair. The authors handle the cross-view domain gap by using polar transform to align the aerial image up to an unknown azimuth angle, which helps the network learn the feature correspondences between polar aerial and ground images instead of having to learn the geometry relation between these two domains. Figure~\ref{fig:Geometrics}(b) shows an overview of the JLO method, where the polar transform is applied to an aerial image. Two stream-CNN networks are applied to extract polar-transformed aerial and ground-level image features. The orientation of the ground-level image concerning the aerial image is estimated using the correlation given the extracted feature volume representation. After that, the aerial features are shifted and cropped to get the image part that probably matches ground features, and the similarity is used for location retrieval. The novelty of this method is that it presents the first image-based geo-localization technique for estimating the orientation and position of ground-level images independent of their field of view. The JLO achieved the highest top-1 accuracy compared with other recent works~\cite{workman2015wide,zhai2017,vo2016localizing,hu2018, Liu_2019_CVPR, regmi2019bridging}, with a value of 99.67\% for CVUSA~\cite{workman2015wide} and 97.32\% for  CVACT~\cite{Liu_2019_CVPR}. 

\vspace{1mm}
\noindent\textbf{LONN*~\cite{Liu_2019_CVPR}:} stands for Lending Orientation to Neural Networks is a ground-level to aerial image matching method based on a Siamese network with explicit orientation encoding of each image pixel. The proposed method has been evaluated using the CVUSA benchmark dataset. In this study, Siamese-type two-branch CNN has been applied, where each branch will learn the useful features for comparing the views. The understanding of an image's orientation facilitates the localization task, and based on that; the authors presented a method for injecting the orientation information for each pixel via orientation maps. Figure~\ref{fig:Geometrics}(c) shows an overview of the LONN~\cite{Liu_2019_CVPR}, where ground-level and aerial images are fed to the Siamese network along with their associated orientation maps. The triplet loss function receives the two learned feature vectors and uses them to train the network. The orientation map injection can be done in two ways: either concatenated with the input layer only or injected in an intermediary convolution layer. The novelty of this work lies in proposing a simple approach for incorporating the orientation information of each pixel. Also, a novel Siamese-based CNN network is used to learn the feature embedding orientation and appearance information. The results showed that the proposed method outperformed other recent works Workman~\textit{et al.}~\cite{workman2015wide}, Zhai~\textit{et~al.} ~\cite{zhai2017}, Vo~\textit{et~al.}~\cite{vo2016localizing}, and Hu~\textit{et~al.} ~\cite{hu2018} with recall top-1 equal to 93.19\%.

\vspace{1mm}
\noindent\textbf{GT-CrossView~\cite{vo2016localizing}:} This model tackled ground-level to satellite image matching by investigating two categories of CNN networks. The first one is representation learning networks for embedding the cross-view images into the same feature space and classification networks for identifying matches. The second category is Siamese and triplet networks for computing similarity. A new Distance-Based Logistic Loss (DBL) function has introduced a novel version of Siamese and triplet networks. In this study, the orientation alignment is considered in addition to the spatial one, so learning the rotation invariant (RI) representation of satellite images helps learn complex representations without using more or larger filters and is done by simply applying random rotation to the training data, i.e., data augmentation. For better learning of representation, the authors proposed an auxiliary loss function where the added rotation amount can be utilized as a label. Figure~\ref{fig:Geometrics}(d) illustrates the network architecture that augments data through random rotation and incorporates an additional branch for orientation regression. The results demonstrate that both classification and hybrid (Siamese classification) networks performed better than feature representation networks, with an accuracy of 90\% and 91.5\%, respectively. Also, Siamese-DBL and triplet-DBL outperformed other original networks with accuracy rates of 88\% and 90.2\%, respectively. The images of a street view panorama are obtained by randomly querying the Google Map of the United States across 11 cities.

\vspace{1mm}
\noindent\textbf{CVM-Net~\cite{hu2018}:} It is an architecture for cross-view ground-to-aerial geolocation, consisting of two subnetworks, as shown in Figure~\ref{fig:Geometrics}(e). The first one, CVM-Net-I, contains two aligned NetVALDs layers that collect the local features into one common space from various views. The second one, CVM-Net-II, consists of two weight-shared NetVALDs layers to convert the local features into a shared space before aggregating to get the global descriptor. CVM-Net~\cite{hu2018} is notable because the authors combine a Siamese network with NetVALDs layers and learn robust cross-view image representations for matching tasks. In addition, the authors proposed a novel loss called weighted soft margin, which enhances retrieval accuracy and makes the training convergence faster. The results illustrate that CVM-Net-I outperforms CVM-Net-II when AlexNet and VGG-16 are employed as local feature extractors; the accuracy reaches 65.4\% and 91.4\%, respectively. Furthermore, CVM-Net-I and CVM-Net-II outperform other existing methods~\cite{zhai2017},~\cite{vo2016localizing},~\cite{workman2015wide}, and the Siamese network of VGG16 and AlexNet baselines, achieving an accuracy of 67.9\% for cropped images and 91.4\% for panorama images evaluated on CVUSA and Dayton~\cite{vo2016localizing}, respectively.

\subsection{Siamese Network-based Models}
These models utilize Siamese or triplet network architectures for image-matching tasks between ground-level and aerial/satellite views. They are focused on learning similarity metrics or embeddings to identify matches between images from different viewpoints, emphasizing the importance of capturing semantic and structural similarities across domains.

\begin{figure}[tbp]
    \centering
    \includegraphics[width=0.99\linewidth]{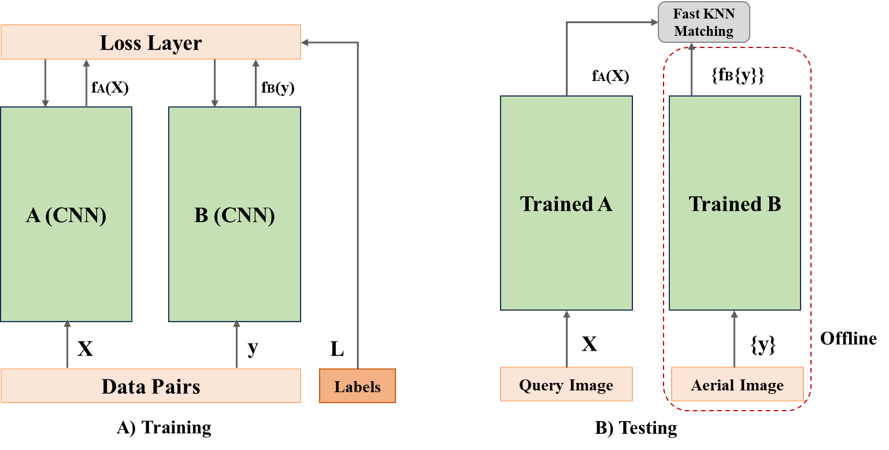}
    \caption{WhereCNN~\cite{Lin_2015_CVPR} architecture overview. }
    \label{fig:Where-CNN}
\end{figure}

\vspace{1mm}
\noindent\textbf{Where-CNN~\cite{Lin_2015_CVPR}:} An early Siamese network~\cite{chopra2005learning} inspired the street-view and aerial image matching technique known as Where-CNN. It consists of two CNNs modified from~\cite{krizhevsky2017imagenet} (see Figure~\ref{fig:Where-CNN}) having different or identical standard parameters. Regarding different parameters, both learn domain-specific deep representations (called Where-CNN-DS). On the other hand, when parameters are shared, both learn a general representation between aerial and street-view images (called Where-CNN). Moreover, the contrastive foss function proposed in~\cite{hadsell2006dimensionality} is utilized and is trained using Caffe~\cite{jia2014caffe}. The features extracted via Where-CNN and Where-CNN-DS outperform Image-CNN, Places-CNN, and hand-crafted features. Further, Where-CNN-DS, a domain-specific model, achieved slightly better performance than Where-CNN, a domain-independent model. The street view and 45-degree aerial images are collected from Google across seven cities.

\subsection{MLP-based Models}
These models are distinguished using Multi-Layer Perceptron (MLP) architectures for aerial-to-street-view generation. MLPs offer flexibility in capturing complex patterns and relationships within the data, making them suitable for tasks like image synthesis.

\begin{figure}[tbp]
    \centering
    \includegraphics[width=0.99\linewidth]{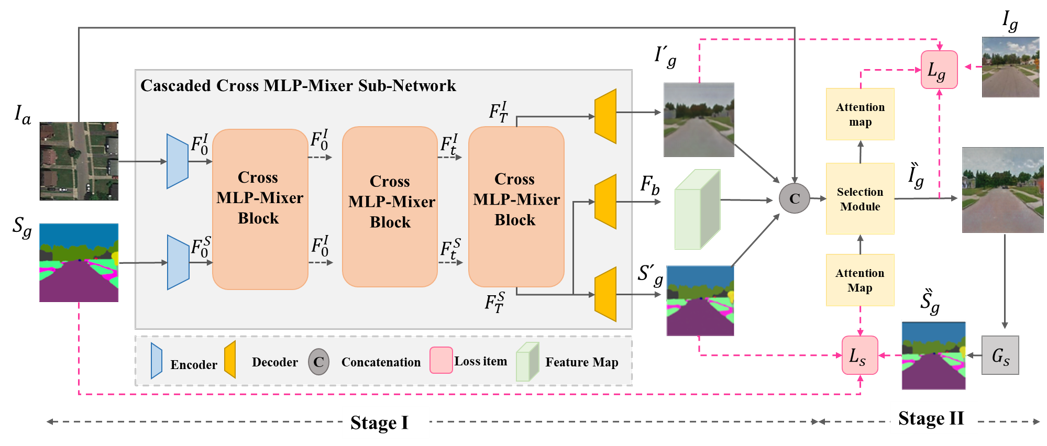}
    \caption{CrossMLP architecture overview taken from~\cite{ren2021cascaded}. }
    \label{fig:CrossMLP}
\end{figure}

\vspace{1mm}
\noindent\textbf{CrossMLP~\cite{ren2021cascaded}:} It is a two-stage framework for aerial-to-street-view generation with a novel cascaded Cross MLP-Mixer (CrossMLP) sub-network as the first stage and a refined pixel loss as the second stage. Figure~\ref{fig:CrossMLP} depicts the overall architecture, wherein the first stage, the Cross-MLP sub-network, stacks multiple Cross-MLP blocks for generating the street view and its corresponding semantic map progressively. Each CrossMLP module captures the structure and appearance information within each view. It can learn the latent correspondence between two inputs by applying a sequential MLP. Then, latent cues guide the transformation from aerial to street view. The refined pixel loss helps with the noisy semantic label by adding more reasonable regularization to the total optimization loss, and with less redundancy, the pixel-level loss becomes more compact. The novelty of this work lies in proposing a two-stage framework for generating street-view images using a novel CrossMLP module. Using CrossMLP to learn the latent mapping cues between aerial-view and street-view semantic maps helps with cross-view translation with accurate appearance and geometry structure. The results showed that CrossMLP performed better than other recent works: Pix2Pix~\cite{isola2017}, X-Fork~\cite{regmi2018}, X-Seq~\cite{regmi2018}, and SelectionGAN~\cite{tang2019multi}, except in terms of PSNR, SD, and SSIM, achieved a bit lower results compared with the others. This study evaluates the method on two benchmark datasets: CVUSA~\cite{workman2015wide} and Dayton~\cite{vo2016localizing}.

\subsection{Recent Advancements}
\begin{figure*}[tbp]
\centering
\begin{tabular}{cc}
\includegraphics[width=\columnwidth]{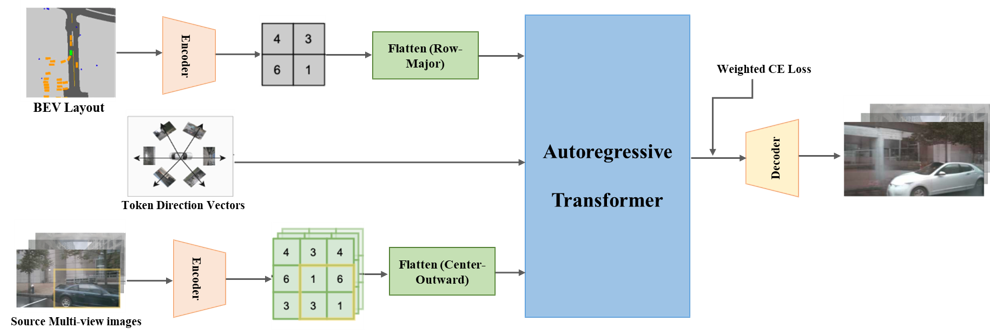}& 
\includegraphics[width=\columnwidth]{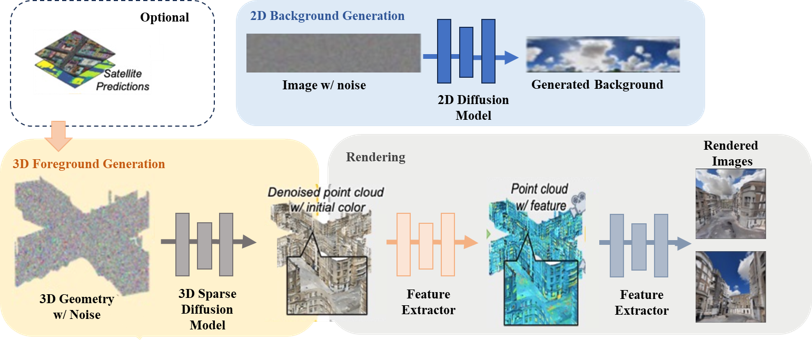}\\
(a) &
(b)\\
 \vspace{1mm}\\

\multicolumn{2}{c}{\includegraphics[width=0.7\textwidth]{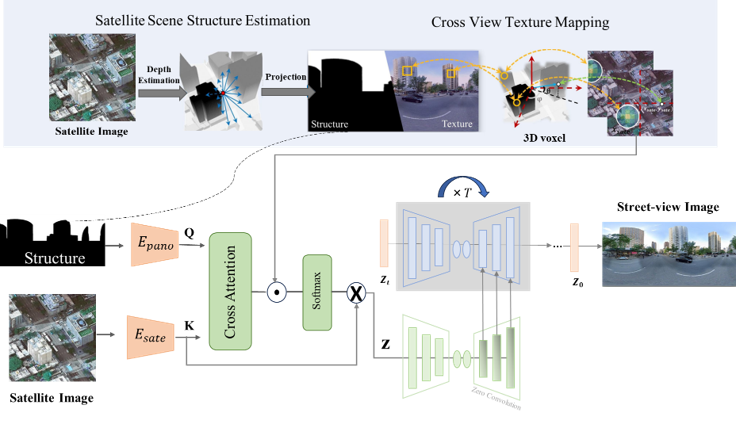}} 
\\
\multicolumn{2}{c}{(c)} \\
\end{tabular}
\caption{Recent Advancements . (a) BevGen \cite{swerdlow2024street}, (b) Sat2Scene~\cite{li2024sat2scene}, (c) CrossViewDiff \cite{li2024crossviewdiff}}
\label{fig:Recent_Advancements}
\end{figure*}

\textcolor{black}{
\vspace{1mm}
\noindent\textbf{BEVGen~\cite{swerdlow2024street}:} It proposed a conditional GAN-based model for generating realistic street-view images from Bird's Eye View layout for traffic scenarios. The evaluation is conducted on the NuScenes 360 driving dataset~\cite{caesar2020nuscenes}. Figure \ref{fig:Recent_Advancements} (a) shows an overview of BEVGen architecture, where it encoded both street-view images and BEV layout to latent discrete representation and flattened them for the autoregressive transformer. Subsequently, the spatial embeddings are added to the street view and Birds Eye View (BEV) tokens, and the pairwise bias is combined with the attention weight. After that, the decoder receives the tokens to get the generated street-view images. BEVGen compares with the baseline model, which has the same GPT architecture and uses identical first-stage encoders and decoders. The result indicates that BEVGen outperformed the baseline on metrics such as Fréchet Inception Distance (FID), road mIoU, and vehicle mIoU. As a limitation, BEVGen~\cite{swerdlow2024street} faces challenges in accurately generating small objects like people and specific vehicles.}


\textcolor{black}{
\noindent\textbf{Sat2Scene \cite{li2024sat2scene}:} Li \textit{et al.} introduces framework for direct 3D scenes video generation given satellite images based on diffusion model. 3D sparse representations have been incorporated with the diffusion model to ensure the images' consistency and enable the generation of street-view features that closely associated with the geometry directly in 3D space. Sat2Scene consists of three steps: generation step, feature extraction step, and rendering step. In the generation step assigns color for the foreground point cloud by utilizing 3D diffusion model combined with the sparse convolutions, while background panorama is generated using a 2D diffusion model. The feature extraction step is responsible for extracting scene features closely associated with the point cloud. The rendering step generates scene frames from arbitrary views using neural rendering. Figure \ref{fig:Recent_Advancements} (b) shows an overview of Sat2Scene architecture. The proposed model was evaluated on two datasets: OmniCity \cite{li2023omnicity} and HoliCity \cite{zhou2020holicity} datasets.
The results demonstrated that, Sat2Scene is able to generate a realistic and texture-consistent street-view image sequences. }

\textcolor{black}{
\noindent\textbf{CrossViewDiff \cite{li2024crossviewdiff}:}
Lin \textit{et al.} proposed a diffusion-based framework for satellite to street-view synthesis task. The CrossViewDiff model presented a novel cross-view texture and satellite scene structure estimation modules, as illustrated in Figure \ref{fig:Recent_Advancements} (c). It leveraging from the imaging and geometric relationships between the satellite and street-view pairs to extract the textual and structural information from the satellite images. These extracted information have been used as a control to guide the street-view generation process, resulting in improved fidelity of the generated street-view images. The authors have used GPT-4o for evaluating the generated street-view images, in addition to the standard evaluation metrics, providing more comprehensive evaluation of the generated street-view images. The proposed model was evaluated on three datasets: CVUSA, CVACT, and OmniCity \cite{li2023omnicity}. The results lustrated that, CrossViewDiff achieves better results with an average increase of $39\%$ in FID, $35.5\%$ GPT-based score, and $9\%$ in SSIM, outperforming other models. 
}

\textcolor{black}{
\noindent After reviewing the existing methods for satellite to street-view synthesis task, we can see that GANs have been widely adopted in this domain. However, GANs may suffer from several issues during the training process, which results in instable training and affect the generated results. The first one is mode collapse which results to a lack of diversity in the generated results. The second one is vanishing gradient issue which arises when the discriminator is very effective in classifying the images into real and fake, which results in training instability and no feedback provided to the generator for improvements. The the third one is non-convergence issue which occurs due to gradient optimizer's failure to reach Nash equilibrium. These issues in during GAN training, resulting in generated images with unrealistic, blurry, and less diversity outputs \cite{saad2024survey}. Nonetheless, the recent advancements, transformers and diffusion models offer significant improvements that help to overcome these limitations. Diffusion models it works by gradually removing Gaussian noise from images to generate an image that closely reflect the target distribution, which improves the training stability and results in the generation of high-quality, more realistic, and finer details. However, diffusion models are computationally expensive, it requires several iterative steps in denoising process, resulting in increasing the time and memory usage which limit their scalability for real-time applications and large datasets. On the other hand, transformers shows strong capabilities in modeling long-range spatial dependencies, enabling them to capture detailed global context, which is considered as important feature for synthesizing detailed environment, such as street-view images \cite{li2022tcgan}. However, it is computationally expensive  and memory intensive which affect its scalability in limited resources environments. }

\begin{figure*}[tbp]
\centering
\begin{tabular}{cc}
\includegraphics[width=\columnwidth]{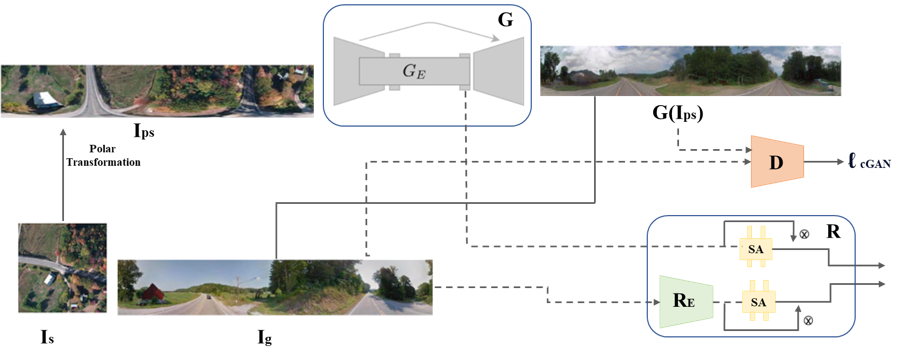}& 
\includegraphics[width=\columnwidth]{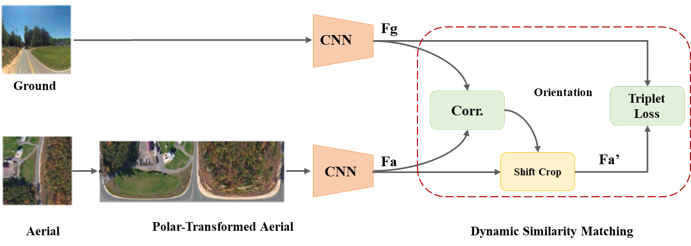}\\
(a) &
(b)\\

\includegraphics[width=0.9\columnwidth]{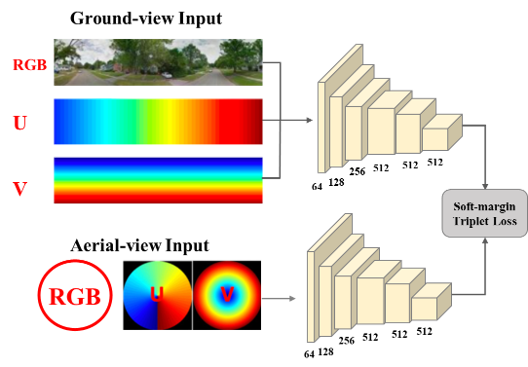}& 
\includegraphics[width=0.9\columnwidth]{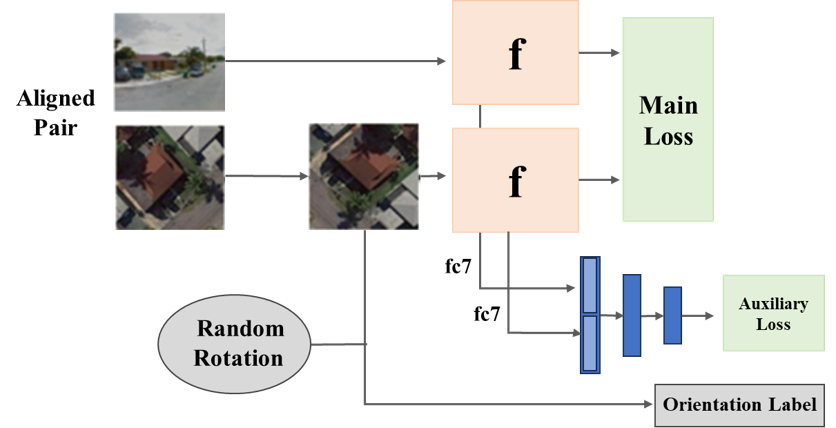}\\
(c) &
(d) \vspace{1mm}\\

\multicolumn{2}{c}{\includegraphics[width=0.7\textwidth]{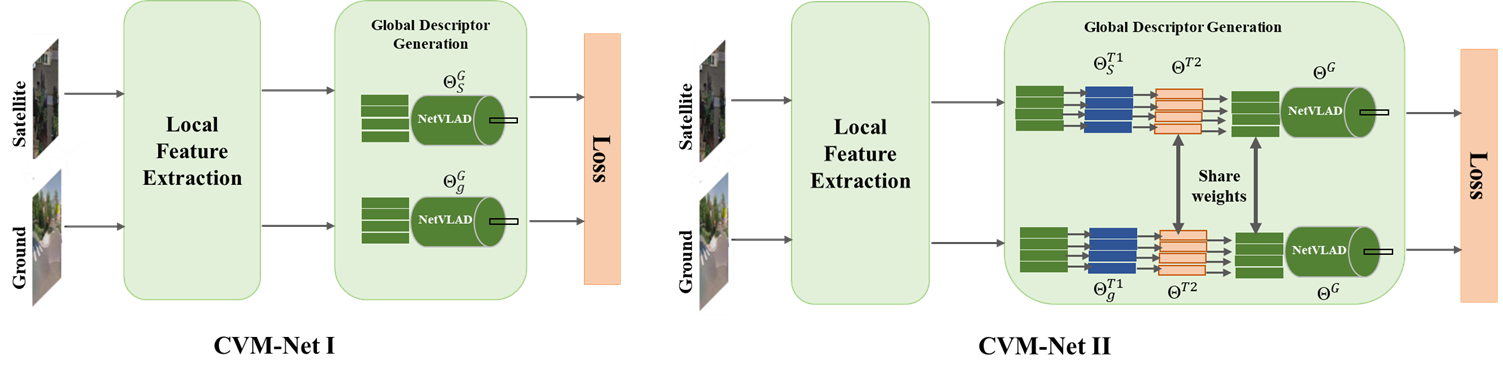}} 
\\
\multicolumn{2}{c}{(e)} \\
\end{tabular}
\caption{Geometric correspondence models. (a) SSVS~\cite{toker2021}, (b) 
JLO~\cite{shi2020looking}, (c) LONN~\cite{Liu_2019_CVPR}, (d)
GT-CrossView~\cite{vo2016localizing}, and (e) CVM-Net~\cite{hu2018}.}
\label{fig:Geometrics}
\end{figure*}

\newpage
\begin{table*}[tbp]
\centering
\resizebox{17cm}{!}{%
\Large
\renewcommand{\arraystretch}{1.0}
\begin{tabular}{p{5cm}lp{3cm}p{4cm}p{5cm}cp{17.5cm}}
\hline
\textbf{Study} & \textbf{Year} & \textbf{Dataset} & \textbf{Method} & \textbf{Task} & \textbf{Metrics} & \textbf{Pros and Cons} \\ \hline
WAGL \cite{workman2015wide}  & 2015 & \begin{tabular}[c]{@{}l@{}}- Charleston\\ - San Francisco\end{tabular}      & - CNN                                                                                    & - Cross-view matching                                                                          & - Accuracy      & Pros:  \newline - Applying satellite features extraction at multiple spatial scales enhances the method's ability to handle perspectives variations, resulting in improved the performance of cross-view matching. \newline Cons: \newline - The multi-scale feature extraction from satellite images introduces computational complexity which may limit its salability for real-world applications with limited resources.                                             \\ \hline
Where-CNN \cite{Lin_2015_CVPR}      & 2015 & - Collected from Google across seven cities.                                   & \begin{tabular}[c]{@{}l@{}}- CNN\\ - Siamese network\end{tabular}                      & - Cross-view matching                                                                          & - Accuracy & Pros: \newline - The method performs geo-localization using only satellite image, eliminating the need for street-view images, which make it applicable in scenarios where the ground-level data unavailable. \newline Cons: \newline - The orientation estimation of the street-view images could be unknown, the system need to test different possible orientation during the testing time to find the correct orientation. This process could be time-consuming and computationally expensive, making it less practical for time-sensitive applications.                                                                                                           \\ \hline
GT-CrossView \cite{vo2016localizing}       & 2016 & - CVACT                                                                       & \begin{tabular}[c]{@{}l@{}}- CNN \\ - Siamese network\\ - Triplet network\end{tabular} & - Cross-view matching                                                                          & \begin{tabular}[c]{@{}l@{}}- Accuracy\\ - Recall\end{tabular}  & Pros: \newline - The explicit prediction of relative rotations of satellite and street-view images improves the retrieval accuracy. \newline Cons: \newline - Sensitive to the quality of rotation predication. Incorrect rotation can negatively impact the retrieval accuracy.

\\ \hline
PGSL \cite{zhai2017}     & 2017 & - CVUSA                                                                       & \begin{tabular}[c]{@{}l@{}}- CNN\\ - GAN\end{tabular}                                  & \begin{tabular}[c]{@{}l@{}}- Cross-view matching \\ - Cross-view synthesis\end{tabular}      & - Average precision & Pros: \newline - The method demonstrates the ability of extract semantic features directly from the satellite images, therefore avoiding the need of manual annotation. \newline Cons: \newline -  The use of multiple sub-networks increases the computational and memory requirements.                                                                                          \\ \hline
X-Seq, X-Fork \cite{regmi2018}    & 2018 & \begin{tabular}[c]{@{}l@{}}- CVUSA\\ - Dayton\end{tabular}                                                              & - cGAN                                                                                   & - Cross-view synthesis                                                                         & \begin{tabular}[c]{@{}l@{}}- Accuracy \\ - Inception score\\ - SSIM\\ - PSNR\\ - Sharpness\\ Difference\end{tabular} & Pros: \newline -  Both X-Fork and X-Seq generate both street-view images and their corresponding semantic segmentation maps, facilitating the learning of more informative features that preserve the street-view consistency. \newline Cons: \newline - the reliance on annotated semantic segmentation data, limits the scalability of the propose methods to datasets that lack such annotations.

\\ \hline
CVM-Net \cite{hu2018}       & 2018 & \begin{tabular}[c]{@{}l@{}}- CVUSA\\ - Dayton\end{tabular}                  & \begin{tabular}[c]{@{}l@{}}- CNN\\ - VLAD descriptor\end{tabular}                      & - Cross-view matching                                                                          & - Recall &           Pros: \newline - Efficient training by utilizing a weighted soft margin loss that accelerate the training convergence. \newline - CVM-Net-I not sharing weights between the two views NetVLAD branches, helps the network to learn view-specific representations, resulting in more effective learning for cross-view matching task. \newline Cons: \newline -  Non-shared NetVLADs restults increase the number of parameters, resulting in higher model complexity.                                                                                                \\ \hline
cGANs \cite{deng2018}     & 2018 & - Collected from Geograph API and Google Map static API                       & - cGAN                                                                                   & \begin{tabular}[c]{@{}l@{}}- Cross-view synthesis\\ - Land-cover classification\end{tabular} & - Accuracy    & Pros: \newline - The method explores and compares different embedding techniques, demonstrating its flexibility and provides insight into how they affect the generated results and help for future enhancements.   \newline Cons: \newline - Despite using dense feature maps, the generated images lack details, limiting the method effectiveness for downstream tasks such as classification.                                                                                                                                           \\ \hline

\end{tabular}}

\caption{Summary of Reviewed Related Studies in the Literature}
\label{tab:summary-LR}
\end{table*}

\newpage

\begin{table*}[tbp]
\centering
\resizebox{17cm}{!}{%
\Large
\renewcommand{\arraystretch}{1.0}
\begin{tabular}{p{5cm}lp{3cm}p{4cm}p{5cm}lp{17.5cm}}
\hline
\textbf{Study} & \textbf{Year} & \textbf{Dataset} & \textbf{Method} & \textbf{Task} & \textbf{Metrics} & \textbf{Pros and Cons} \\ \hline
BridgeGAN \cite{zhu2018generative}      & 2019 & - Collected by {[}23{]}                                                       & - GAN                                                                                    & - Cross-view synthesis                                                                         & \begin{tabular}[c]{@{}l@{}}- SSIM \\ - PSNR\\ - LPIPS\end{tabular}       &                               Pros: \newline - Utilizing homograph view works as bridge between the frontal and bird-eye views, enabling model to learn the transformation gradually. \newline Cons: \newline - The use of three distinct GANS for separate domain frontal, homography and bird-eye views, with shared layers, makes the model computationally expensive during training phase.            \\ \hline

SelectionGAN \cite{tang2019multi}     & 2019 & \begin{tabular}[c]{@{}l@{}}- Dayton\\ - CVUSA\\ - Ego2Top\end{tabular}      & - cGAN                                                                                   & - Cross-view synthesis                                                                         & \begin{tabular}[c]{@{}l@{}}- Accuracy\\ - Inception score\\ - SSIM\\ - PSNR\\ - SD\end{tabular}                & Pros: \newline - Multi-channel attention module enhances the learning effectiveness. \newline - Cascaded semantic guidance generation, allow  the model to generate detailed images by progressively refining the initial generation. \newline Cons: \newline - The multi-channel attention and multi intermediate generation introduces more computational cost.    \\ \hline

FeatureFusion \cite{regmi2019bridging}    & 2019 & - CVUSA                                                                     & - cGAN                                                                                   & \begin{tabular}[c]{@{}l@{}}- Cross-view matching \\ - Cross-view synthesis\end{tabular}      & - Accuracy & Pros: \newline 
- Incorporating edge maps and semantic segmentation provides additional boundary information. \newline 
- Applying feature fusion strategy helps to integrate features from different source, resulting in enhancing the feature learning. \newline - Multi-resolution aggregation from multiple convolution layers allow the model to preserve global context to local details representations. \newline Cons: \newline - Feature fusion and multi-resolution aggregation introduce computational complexity. 
\\ \hline
LONN \cite{Liu_2019_CVPR}      & 2019 & \begin{tabular}[c]{@{}l@{}}- CVUSA\\ - CVACT\end{tabular}                   & \begin{tabular}[c]{@{}l@{}}- CNN\\ - Siamese network\end{tabular}                      & - Cross-view matching                                                                          & - Recall                                                                             & Pros: \newline - Incorporating per-pixel orientation information in a straightforward manner through orientation maps, allow to inject the geometric information without increasing model complexity. \newline - The simple design of the model contributes to facilitating model training, debug and deployment, improving its usability for real-world applications. \newline   Cons: \newline - The simplicity of this design limits its scalability to more complex geospatial tasks, which require more deep feature modeling.     \\ \hline
LGGAN \cite{Tang_2020_CVPR}     & 2020 & \begin{tabular}[c]{@{}l@{}}- Dayton\\ - CVUSA\end{tabular}                  & - GAN                                                                                    & - Cross-view synthesis                                                                         & \begin{tabular}[c]{@{}l@{}}- Inception score\\ - Accuracy\\ - SSIM\\ - PSNR\\ - KL\\ - SD\end{tabular}     & Pros: \newline -  Semantic guided generation, which ensures the structural consistency of the generated images.  \newline - Combines global and local class-specific generation, enabling the model to maintain the overall scene coherence and capture local details. \newline Cons: \newline - Multiple interconnected components increased the overall model complexity, making training more challenging.           \\ \hline
GeoUrban \cite{lu2020}       & 2020 & - Collected from Google Street-view using Google API                          & - GAN                                                                                    & - Cross-view synthesis                                                                      & \begin{tabular}[c]{@{}l@{}}- Accuracy\\ - PSNR\\ - SSIM\\ - SD\\ - mIoU\end{tabular}   & Pros: \newline -  Employ pre-processing step to mitigate the image misalignment by calculating overlap ratio. \newline Cons: \newline -  Does not correct the misaligned data, it just filter out images with low ratio which may affect the dataset diversity and model robustness.                \\ \hline
JLO \cite{shi2020looking}    & 2020 & \begin{tabular}[c]{@{}l@{}}- CVUSA\\ - CVACT\end{tabular}                   & \begin{tabular}[c]{@{}l@{}}- CNN\\ - Dynamic\\ similarity matching\end{tabular}          & - Cross-view matching                                                                          & - Recall   & Pros: \newline - Robust to image misorientation and narrow Field of View (FoV). \newline - Effective orientation estimation using azimuth angle. \newline Cons: \newline - Azimuth angle sensitivity, where small misalignment affect the localization accuracy.                                                                                             \\ \hline

\end{tabular}}

\caption*{\textbf{Table 1:} Summary of Reviewed Related Studies in the Literature (continued)}
\end{table*}

\newpage

\begin{table*}[tbp]
\centering
\resizebox{17cm}{!}{%
\Large
\renewcommand{\arraystretch}{1.0}
\begin{tabular}{p{5cm}lp{3cm}p{4cm}p{5cm}lp{17.5cm}}
\hline
\textbf{Study} & \textbf{Year} & \textbf{Dataset} & \textbf{Method} & \textbf{Task} & \textbf{Metrics} & \textbf{Pros and Cons} \\ \hline

SSVS \cite{toker2021}    & 2021 & \begin{tabular}[c]{@{}l@{}}- CVUSA\\ - CVACT\end{tabular}                   & \begin{tabular}[c]{@{}l@{}}- cGAN\\ - ResNet34\end{tabular}                            & \begin{tabular}[c]{@{}l@{}}- Cross-view matching \\ - Cross-view synthesis\end{tabular}      & \begin{tabular}[c]{@{}l@{}}- Recall\\ - SSIM\\ - PSNR\\ - SD\\ - KL-Scores\\ - LPIPS\end{tabular}   & Pros: Multi-tasking framework allow shared features learning. \newline Cons: Join learning increases the computational complexity.                \\ \hline
CrossMLP \cite{ren2021cascaded}      & 2021 & \begin{tabular}[c]{@{}l@{}}- Dayton\\ - CVUSA\end{tabular}                  & - GAN                                                                                    & - Cross-view synthesis                                                                        & \begin{tabular}[c]{@{}l@{}}- Accuracy\\ - Inception score\\ - SSIM\\ - PSNR\\ - SD\\ - KL\end{tabular} &  Pros: \newline - Addresses the noise in the input semantic segmentation, improving the model robustness against noise. \newline Cons: \newline -  The multi-stage architecture, along with components like feature fusion and Cross-MLP blocks, introduces architectural complexity and resource-intensive model. 
\\ \hline
SSVG \cite{shi2022}     & 2022 & \begin{tabular}[c]{@{}l@{}}- CVUSA\\ - CVACT\end{tabular}                   & -  cGAN                                                                                   & - 
 Cross-view synthesis                                                                         & \begin{tabular}[c]{@{}l@{}}- Accuracy\\ - RMSE\\ - SSIM\\ - PSNR\\ - SD\\ - mIoU\end{tabular}                 & Pros: \newline - Applies implicit height map supervision leveraging two-view geometric constraints, thus avoiding the need for annotated data, which may not always available.  \newline Cons: \newline - Sensitive to height estimation errors, leading to large difference between the generated image and its corresponding ground truth when estimating is incorrect.    \\ \hline
PAGAN \cite{wu2022cross}       & 2022 & \begin{tabular}[c]{@{}l@{}}- CVUSA\\ - Orlando-\\Pittsburgh\\ (OP)\end{tabular} & cGAN                                                                                    & Cross-view synthesis                                                                         & \begin{tabular}[c]{@{}l@{}}- Accuracy\\ - SSIM\\ - PSNR\\ - SD\\ - KL \\ - FID\end{tabular}     &         Pros:\newline - Cross-stage Attention module provide efficient feature fusion by using lightweight attention mechanism. \newline - Orientation-aware data augmentation, making the model robust for views changes. \newline Cons:\newline - The lightweight attention may limit the model capabilities in capturing complex features.              \\ \hline

PanoGAN\cite{wu2022cross}                                                    & 2022          & \begin{tabular}[c]{@{}l@{}}- CVUSA\\ - Orlando-\\Pittsburgh\\ (OP)\end{tabular} & - cGAN                                                                                   & - Cross-view synthesis                                                                         & \begin{tabular}[c]{@{}l@{}}- Inception Score\\ - Accuracy\\ - SSIM\\ - PSNR\\ - SD\\ - KL\end{tabular}        & Pros:\newline - Iterative image refinement. \newline - Effective multi-scale feature representation that capture both high-level and low-level information with minimal computation cost. \newline Cons:\newline - Resources-intensive due to multiple components.\newline - Architecture complexity. \newline - Dependency on labeled data, segmentation map, which may not always available. \\ \hline
BevGen\cite{swerdlow2024street} & 2024 & - NuScenes                                                                    & - Autoregressive Transformer                                                             & - Cross-view synthesis                                                                         & - FID scores     & Pros:\newline - Multi-view generation. \newline - Spatial consistency modeling. \newline Cons:\newline   - Multi-view generation increases model complexity.  
 \\ \hline

CrossViewDiff \cite{li2024crossviewdiff} & 2024 & \begin{tabular}[c]{@{}l@{}}- CVUSA\\ - CVACT\\ - OmniCity \end{tabular} & - Diffusion model & - Cross-view synthesis & \begin{tabular}[c]{@{}l@{}}- SSIM\\ - PSNR\\ - SD\\ - FID \\ - KID\end{tabular} & Pros: \newline - Incorporating structure-aware and texture mapping modules, improving the layout consistency and accurate appearance for the generated street-view images.  \newline Cons: \newline - Integrating diffusion model with additional guidance modules introduces high computational costs, as diffusion by is inherently resource-intensive by nature.   

\\ \hline

Sat2Scene \cite{li2024sat2scene} & 2024 & \begin{tabular}[c]{@{}l@{}}- OmniCity\\ - HoliCity \end{tabular} & - Diffusion model & - Cross-view synthesis & \begin{tabular}[c]{@{}l@{}}- SSIM\\ - PSNR\\ - LPIPS\\ - FID \\ - KID \\ - FVD \\ - KVD \end{tabular} & Pros: \newline - 3D street-view generation, is valuable features for real-world applications, like urban planning simulation. \newline Cons: \newline
- Combining diffusion models with 3D sparse representations, may introduce computational and memory cost, especially with large dataset.

\\ \hline

\end{tabular}}

\caption*{\textbf{Table 1:} Summary of Reviewed Related Studies in the Literature (continued)}
\label{tab:summary-LR}
\end{table*}

\newpage
\section{Datasets}\label{sec3}

We discussed the primary three datasets used for satellite-to-street translation as given in Table~\ref{tab:datasetsummary}.

\begin{table*}[t]
    \caption{Summary of the datasets used for Satellite to Street view translation.}

    \centering
    \begin{tabular}{c|c|c|c|cc|ccc}\hline
       &       &Ground-View  &         & \multicolumn{2}{c|}{Resolution}   & \multicolumn{3}{c}{Images}\\    
Dataset& Year  &FoV          & GPS-tag & Images         &Satellite        &Training &Testing &Total\\ \hline\hline
Dayton & 2016  &  -          &  -      &354$\times$354  &354$\times$354   &55,000   &21,048  &76,048      \\  
CVUSA  & 2017  &360          & No      &1232$\times$224 &750$\times$750   &35,532   &8,884   &44,416\\
CVACT  & 2019  &360          & Yes     &1664$\times$832 &1200$\times$1200 &35,532   &92,802  &128,334 \\\hline
    \end{tabular}
    
    \label{tab:datasetsummary}
\end{table*}

\begin{figure*}[h!]
\centering
\hspace{-2.1cm}
 \resizebox{18cm}{!}{
\begin{tabular}{ccc}
\multirow{-5}{*}{Satellite Image}& 
\includegraphics[width=3cm,height=3cm]{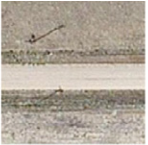} &
\includegraphics[width=3cm, height=3cm]{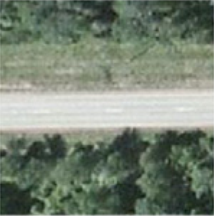}\\

\multirow{-5}{*}{Ground truth} & \includegraphics[width=10cm,height=3cm]{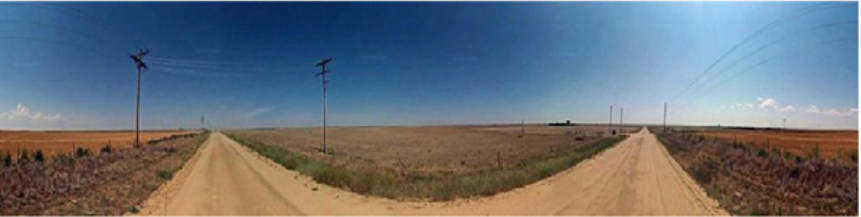}&
\includegraphics[width=10cm,height=3cm]{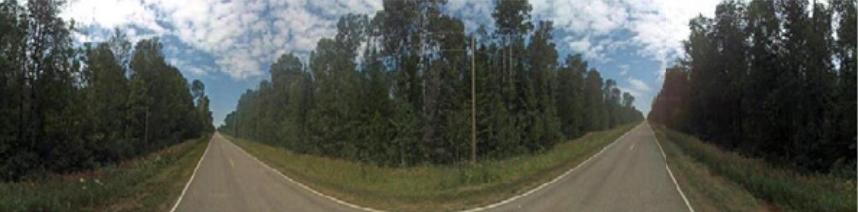}\\

\multirow{-5}{*}{Pix2pix \cite{isola2017}}&
\includegraphics[width=10cm,height=3cm]{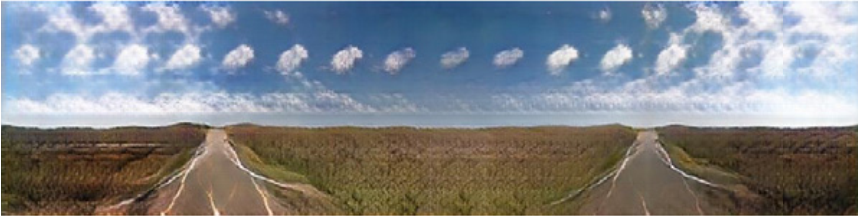}&
\includegraphics[width=10cm,height=3cm]{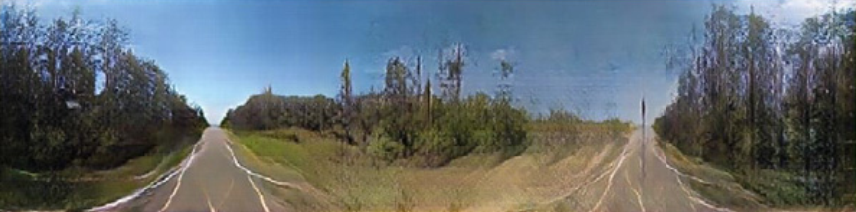}\\

\multirow{-5}{*}{X-Fork \cite{regmi2018}} & 
\includegraphics[width=10cm,height=3cm]{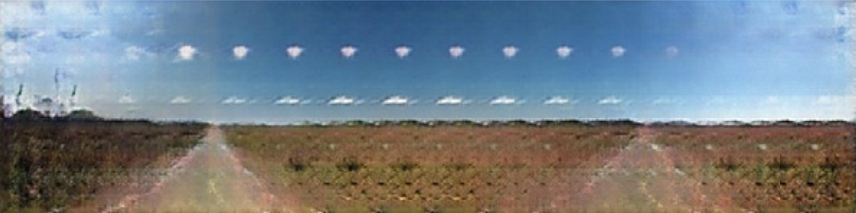}
&\includegraphics[width=10cm,height=3cm]{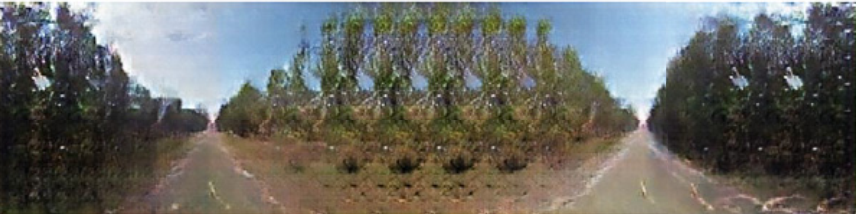}\\

\multirow{-5}{*}{X-Seq \cite{regmi2018}} & 
\includegraphics[width=10cm,height=3cm]{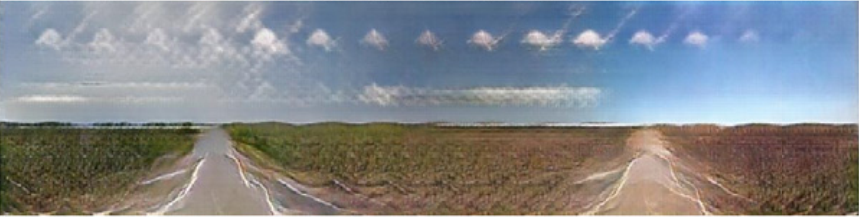}&
\includegraphics[width=10cm,height=3cm]{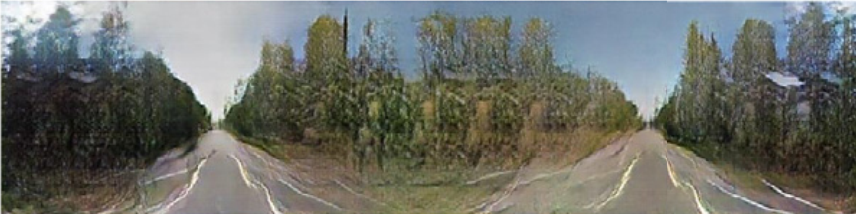}\\

\multirow{-5}{*}{SelectionGAN \cite{tang2019multi}} & 
\includegraphics[width=10cm,height=3cm]{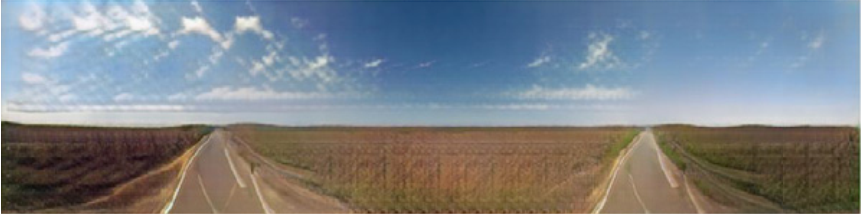} &
\includegraphics[width=10cm,height=3cm]{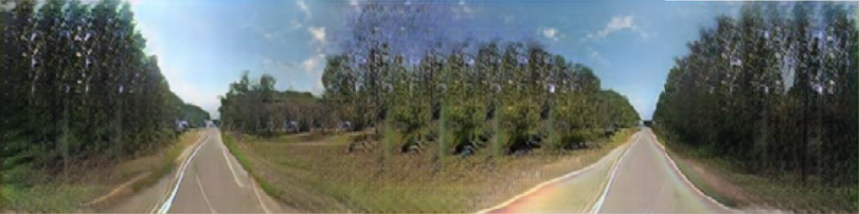}\\

\multirow{-5}{*}{GeoUrban \cite{lu2020}} & 
\includegraphics[width=10cm,height=3cm]{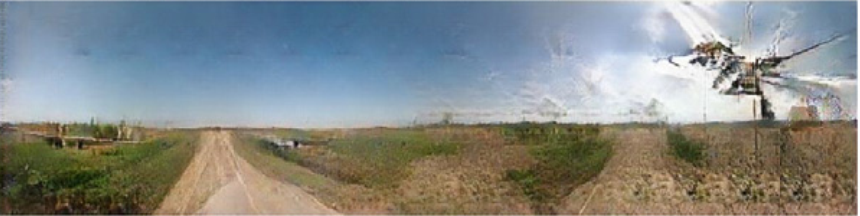}&
\includegraphics[width=10cm,height=3cm]{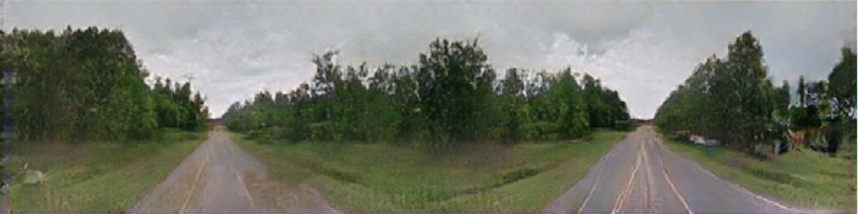}\\

\multirow{-5}{*}{PAGAN \cite{wu2022cross}} & 
\includegraphics[width=10cm,height=3cm]{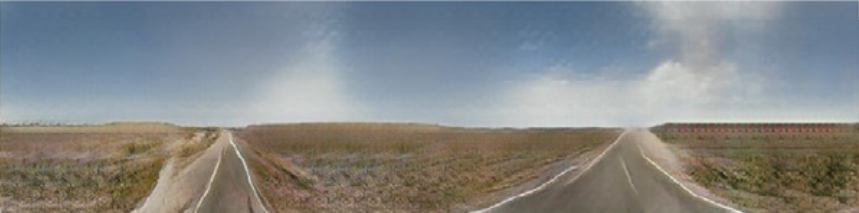}&

\includegraphics[width=10cm,height=3cm]{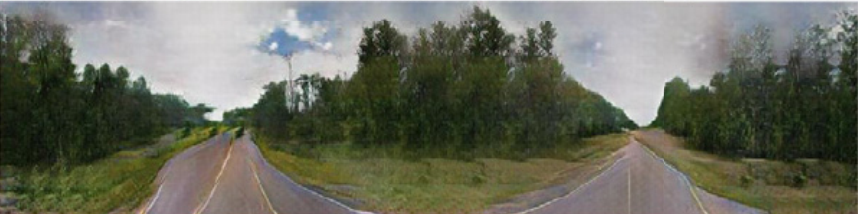}\\

\end{tabular}}
\caption{Qualitative comparison on CVUSA dataset.}
\label{fig:qualitative_cvusa}
\end{figure*}


\begin{figure*}[tbp]
\centering
\hspace{-2.1cm}
 \resizebox{18cm}{!}{
\begin{tabular}{ccc}
\multirow{-5}{*}{Satellite Image}& 
\includegraphics[width=3cm,height=3cm]{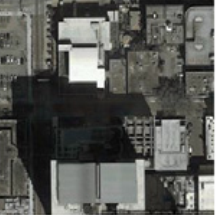} &
\includegraphics[width=3cm,height=3cm]{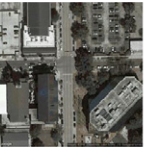}\\

\multirow{-5}{*}{Ground truth} & \includegraphics[width=10cm,height=3cm]{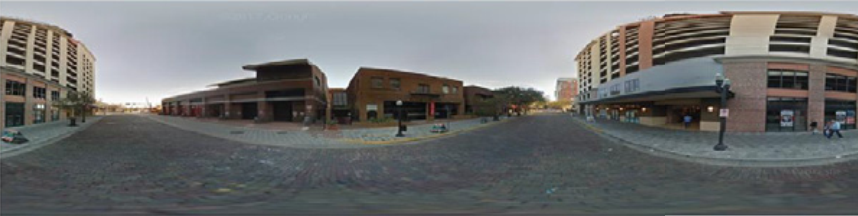} &
\includegraphics[width=10cm,height=3cm]{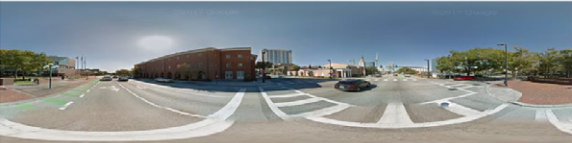}\\

\multirow{-5}{*}{Pix2pix\cite{isola2017}} & \includegraphics[width=10cm,height=3cm]{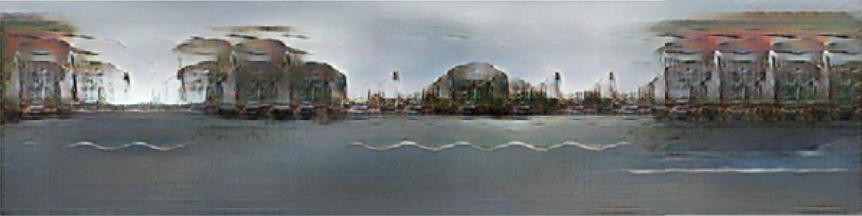}&
\includegraphics[width=10cm,height=3cm]{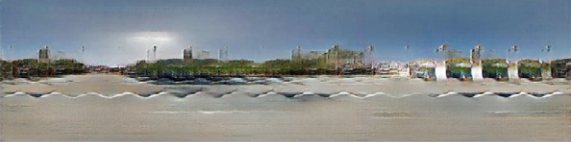}\\

\multirow{-5}{*}{X-Frok \cite{regmi2018}} & \includegraphics[width=10cm,height=3cm]{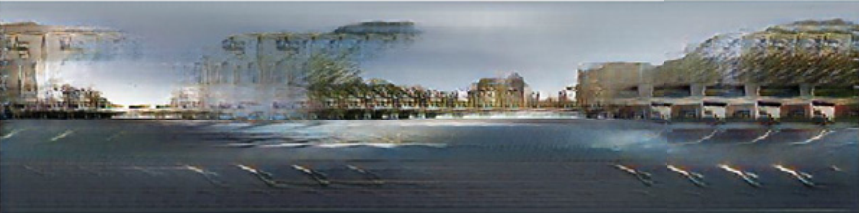}&
\includegraphics[width=10cm,height=3cm]{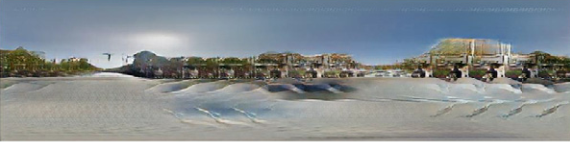}\\

\multirow{-5}{*}{X-Seq \cite{regmi2018}} & \includegraphics[width=10cm,height=3cm]{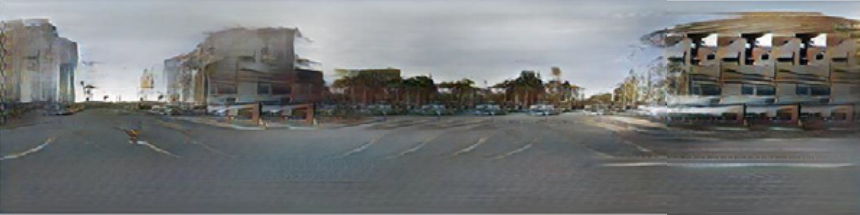}&
\includegraphics[width=10cm,height=3cm]{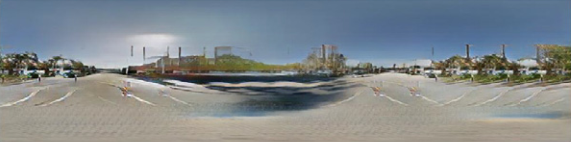}\\

\multirow{-5}{*}{SelectionGAN \cite{tang2019multi}} & \includegraphics[width=10cm,height=3cm]{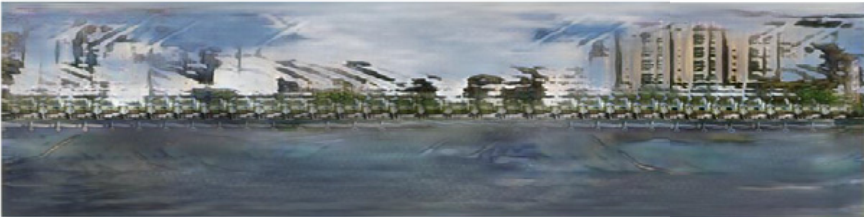}&
\includegraphics[width=10cm,height=3cm]{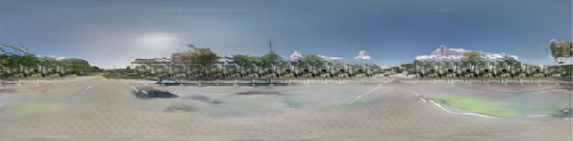}\\

\multirow{-5}{*}{GeoUrban \cite{lu2020}} & \includegraphics[width=10cm,height=3cm]{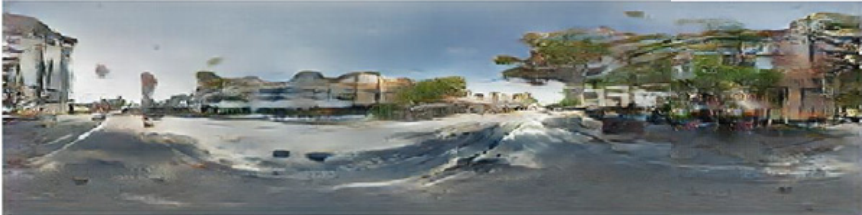}&
\includegraphics[width=10cm,height=3cm]{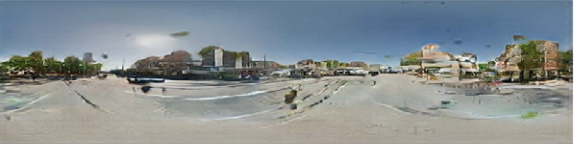}\\ 

\multirow{-5}{*}{PAGAN \cite{wu2022cross}} & \includegraphics[width=10cm,height=3cm]{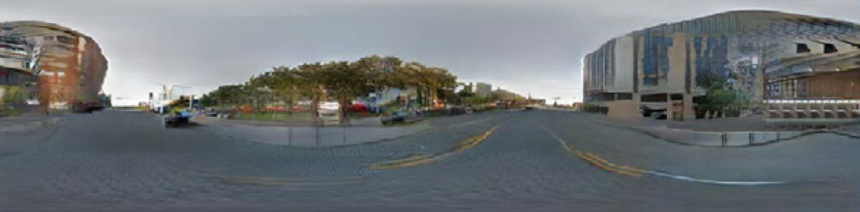}&
\includegraphics[width=10cm,height=3cm]{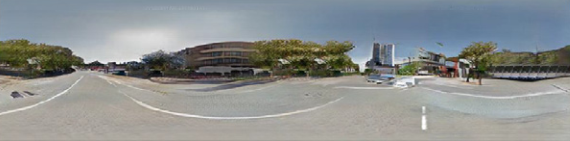}\\

\end{tabular}}
\caption{Qualitative comparison on OP dataset.}
\label{fig:qualitative_op}
\end{figure*}


\begin{figure*}[tbp]
\centering
\hspace{-2.1cm}
 \resizebox{18cm}{!}{
\begin{tabular}{ccc}
\multirow{-5}{*}{Satellite Image}& 
\includegraphics[width=3cm,height=3cm]{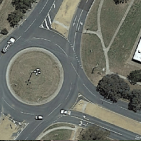} &
\includegraphics[width=3cm,height=3cm]{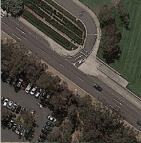}\\

\multirow{-5}{*}{Ground truth} & \includegraphics[width=10cm,height=3cm]{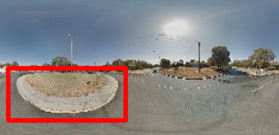} &
\includegraphics[width=10cm,height=3cm]{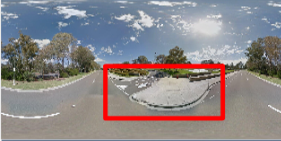}\\

\multirow{-5}{*}{Pix2pix \cite{isola2017}} & \includegraphics[width=10cm,height=3cm]{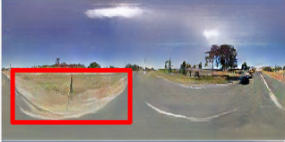}&
\includegraphics[width=10cm,height=3cm]{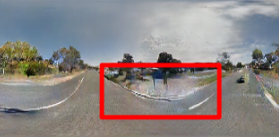}\\

\multirow{-5}{*}{X-Frok \cite{regmi2018}} & \includegraphics[width=10cm,height=3cm]{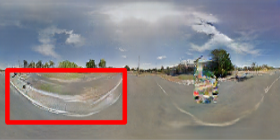}&
\includegraphics[width=10cm,height=3cm]{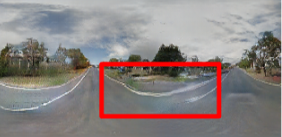}\\

\multirow{-5}{*}{SSVG \cite{shi2022}} & \includegraphics[width=10cm,height=3cm]{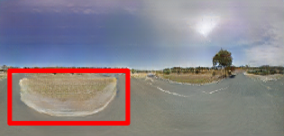}&
\includegraphics[width=10cm,height=3cm]{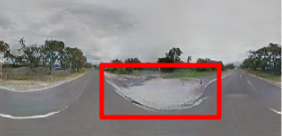}\\

\end{tabular}}
\caption{Qualitative comparison on CVACT dataset.}
\label{fig:qualitative_cvact1}
\end{figure*}

\begin{figure*}[tbp]
\centering
\hspace{-2cm}
 \resizebox{18cm}{!}{
\begin{tabular}{cccc}
Satellite Image& 
ControlNet \cite{zhang2023adding} &
CrossViewDiff \cite{li2024crossviewdiff}&
Ground truth
\\
\includegraphics[width=3cm,height=3cm]{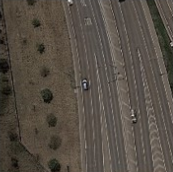} &
\includegraphics[width=10cm,height=3cm]{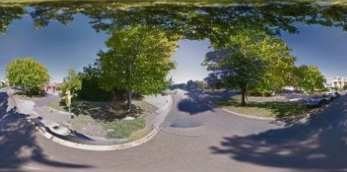} &
\includegraphics[width=10cm,height=3cm]{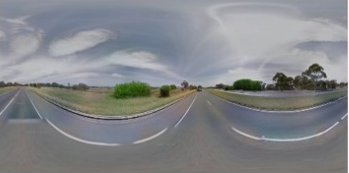}& 
\includegraphics[width=10cm,height=3cm]{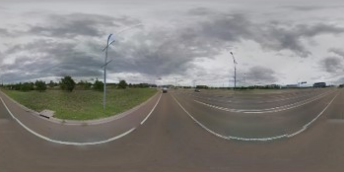}\\
\includegraphics[width=3cm,height=3cm]{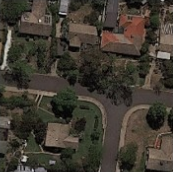}&
\includegraphics[width=10cm,height=3cm]{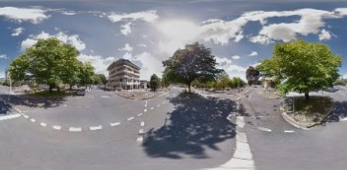}& 
\includegraphics[width=10cm,height=3cm]{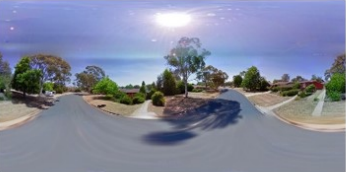}&
\includegraphics[width=10cm,height=3cm]{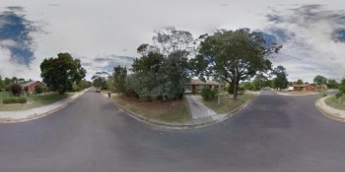}\\
\includegraphics[width=3cm,height=3cm]{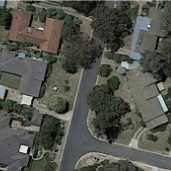} &
\includegraphics[width=10cm,height=3cm]{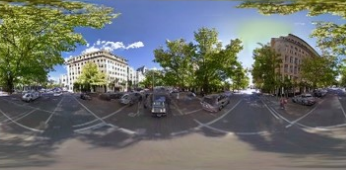}&
\includegraphics[width=10cm,height=3cm]{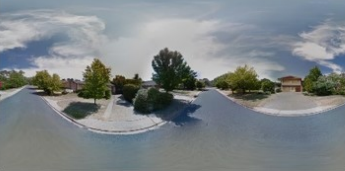}&
\includegraphics[width=10cm,height=3cm]{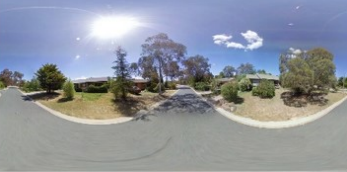}

\end{tabular}}
\caption{Qualitative comparison on CVACT dataset.}
\label{fig:qualitative_cvact2}
\end{figure*}

\begin{figure*}[tbp]
  \centering
  \begin{tabular}{ccc}
    \includegraphics[width=0.32\textwidth]{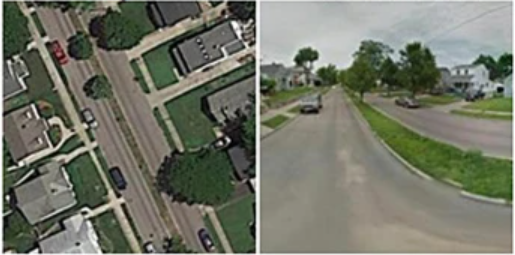}& 
    \includegraphics[width=0.32\textwidth]{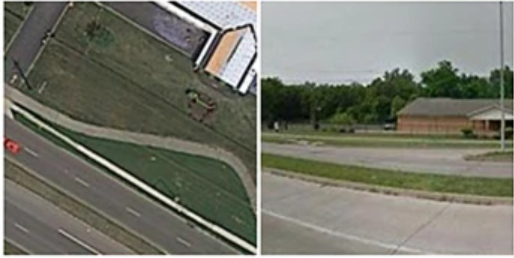}&
    \includegraphics[width=0.32\textwidth]{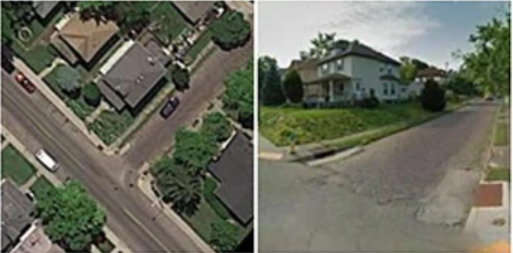}\\
\end{tabular}
\caption{Sample street-view and satellite image pair from Dayton dataset.}
   \label{fig:Dayton}
\end{figure*}

\vspace{2mm}
\noindent\textbf{Dayton} dataset~\cite{vo2016localizing} is introduced for ground-to-aerial or aerial-to-ground image translation or cross-view image synthesis, consisting of 76,048 pairs of satellite and street-view images across 11 different cities in the USA, where 55,000 are reserved for training and the rest for testing. Street-view images are collected from Google Maps of the USA, then several crops are made for each street-view panorama, and for each crop, a satellite image is queried using Google Maps with a resolution of 354$\times$354. Figure~\ref{fig:Dayton} showcases sample images from the dataset.

\begin{figure*}[tbp]
  \centering
  \begin{tabular}{cc}
    \includegraphics[height= 2cm, width=7.8cm]{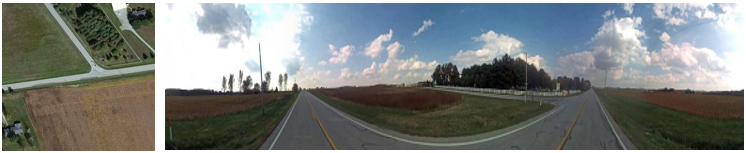}& 
    \includegraphics[height= 2cm, width=7.8cm]{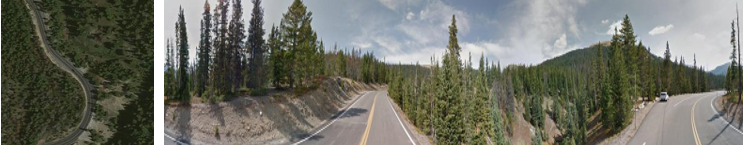}\\
    \includegraphics[height= 2cm, width=7.8cm]{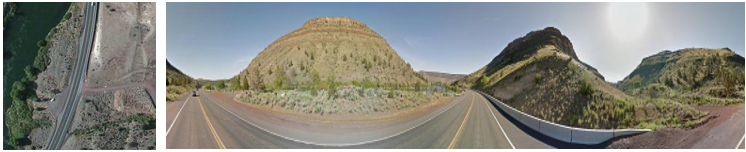}&
    \includegraphics[height= 2cm, width=7.8cm]{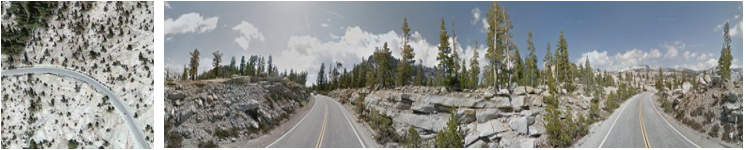} \\
  \end{tabular}
    \caption{Sample street-view and satellite image pair from CVUSA dataset.}
   \label{fig:CVUSA}
\end{figure*}

\vspace{1mm}
\noindent\textbf{CVUSA~\cite{workman2015wide}}
 dataset is an abbreviation of the Crossview United States of America for cross-view geo-localization tasks. The dataset comprises 44,416 satellite and panorama street-view images across the USA, with 35,532 pairs for training and 8,884 for testing. The images are collected from Flicker and Google Street View, having a resolution of 1232 $\times$ 224. For Flicker, the area has been divided into 100$\times$100 grid; 150 images from each grid cell have been downloaded from 2012 onwards, while the pictures for Google Street View are sampled from randomly selected places within the USA. Moreover, the satellite images are obtained from Bing Maps with a resolution of 750 $\times$ 750, where for each street-view image, an 800$\times$800 satellite image is downloaded centered on the exact location. Figure~\ref{fig:CVUSA} showcases sample images from the dataset.

\begin{figure*}[tbp]
  \centering
  \begin{tabular}{cc}
    \includegraphics[height= 2cm, width=7.75cm]{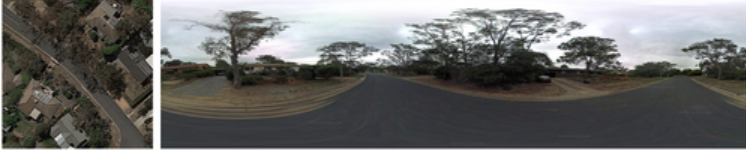} & 
    \includegraphics[height= 2cm, width=7.75cm]{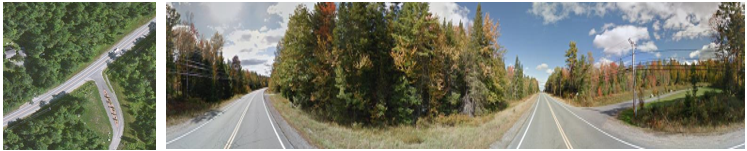}\\ \includegraphics[height= 2cm, width=7.75cm]{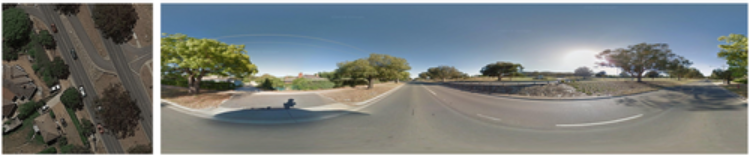} &
    \includegraphics[height= 2cm, width=7.75cm]{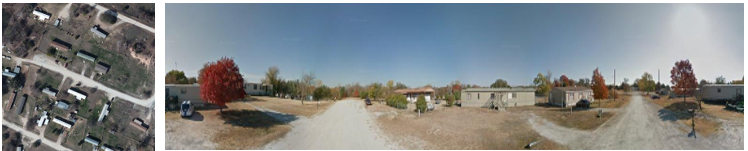}\\
  \end{tabular}
    \caption{Sample street-view and satellite image pair from CVACT dataset.}
   \label{fig:CVACT}
\end{figure*}

\vspace{1mm}\noindent\textbf{CVACT~\cite{Liu_2019_CVPR}}
dataset stands for Australian Capital Territory City-Scale Cross-View, consisting of 128,334 pairs of satellite and panorama street-view images across Canberra City in Australia. The dataset has 35,532 for training and 92,802 for testing. Street-view images have been collected from the Google Street View API with an image resolution of 1664$\times$832. The authors have obtained the satellite images from the Google Map API. Further, the matched satellite image at the GPS position has been downloaded for each street-view image with a resolution of 1200$\times$1200. Figure~\ref{fig:CVACT} showcases sample images from the dataset.

\section{Evaluation Metrics}\label{sec4}
This section discusses the standard evaluation metrics used in literature for evaluating satellite to street-view generation.

\begin{itemize}

\item \textbf{Inception Score (IS):} measures the diversity and quality of the generated images. A good model should produce a reasonable and diverse set of images~\cite{baraheem2023image}. The Inception Score  is defined as  
\begin{equation}
\label{eq:1}
IS= exp( E_{x} DKL (p(y \mid x)  \parallel p(y)))
\end{equation}
Table~\ref{tab:quantitative-comparison-cvusa} shows a few comparative results (i.e., accuracy and inception scores) using Conditional GANs on the CVUSA dataset.

\item \textbf{Fŕechet Inception Distance (FID):} is based on the extracted visual features and calculates the distance between the actual $p_{real}(x)$ and the generated distribution $p _{g(x)}$~\cite{baraheem2023image}. As illustrated below, FID computes the multivariate Gaussian for real and generated images.

\begin{equation}
\label{eq:2}
\begin{aligned}
&FID ( p_{real},   p_{g} ) =  d^{2} (( m_{real},   m_{g}), (  m_{g},  c_{g} ) \\
&=  \parallel  m_{real}, -  m_{g}  \parallel ^{2}  + T_{r} ( c_{real} +   c_{g} - 2 (  c_{real}\ c_{g} )) ^{1/2}  
\end{aligned}
\end{equation}

\item \textbf{Peak Signal to Noise Ratio (PSNR):} is the ratio between maximum signal power to the distortion noise that affects the signal’s ability to be represented accurately~\cite{sara2019image}. In this case, the ratio between two images is calculated in decibels as expressed.  

\begin{equation}
PSNR = 10 \log_{10} \frac{peakval^{2}}{MSE}
\label{eq:PSNR}
\end{equation}

\item \textbf{Structure Similarity Index Method (SSIM):} determines the similarity between the original image and its recovered version. In SSIM, the degradation of images is considered an alteration in how structural information is perceived. Additionally, it collaborates with other key perception-based facts like contrast masking (contrast masking is a process where the degree of distortion is less apparent in the image’s texture) and luminance masking (a process where distortion is less clear at image boundaries). The SSIM is expressed in terms of contrast, luminance, and structure as follows

\begin{equation}
SSIM (x,y) = [l(x,y)]^{ \alpha } \cdot [c(x,y)]^{\beta }  \cdot  [s(x,y)]^{ \gamma }.
\label{eq:ssim}
\end{equation}

The luminance, contrast and structure are expressed as

\begin{equation}
\begin{aligned}
l (x,y) &= \frac{2\mu_{x}\mu_{y}+C_{1}}{ \mu^{2}_{x} + \mu^{2}_{y}  +C_{1}},\\
c (x,y) &= \frac{2 \sigma_{x} \sigma_{y} +C_{2}}{\sigma^{2}_{x} +\sigma ^{2}_{y} +C_{2}},\\ 
s (x,y) &= \frac{\sigma_{xy}+C_{3}}{ \sigma_{x}\sigma_{y}+C_{3}}.
\end{aligned}
\label{eq:ssim_lcc}
\end{equation}

\item \textbf{Kullback–Leibler (KL) divergence:} estimates the similarity of the probability distribution between the real and generated images~\cite{betzalel2022study}. The zero KL value refers to an identical distribution, while the higher KL value refers to less similarity~\cite{solanki2021generative} between the images. The KL divergence can be calculated as follows.

\begin{equation}
KL (p \parallel q)= E_{x} [log({\frac{p(x)}{q(x) }})].
\label{eq:KL}
\end{equation}

\end{itemize}

\section{Qualitative Comparison}
\textcolor{black}{
Figure \ref{fig:qualitative_cvusa} (a) shows a qualitative comparison on the CVUSA dataset. It can be observed that Pix2pix and X-Seq trying to generate images that closely mimic the outlines of the ground truth image. In contrast,  X-Fork and GeoUrban generate images with more artifacts compared to Pix2pix and X-Seq. Also, SelectionGAN demonstrates improvements in generating more clearly shaped streets and street lines, while the PAGAN method generates the most realistic street-view images compered to others. On the other hand, Figure \ref{fig:qualitative_cvusa}(b) shows another qualitative example, clearly showing that X-Fork method generates low-quality images with noticeable artifacts. The Pix2pix and X-Seq methods generate better street-view images compared to X-Fork; however, artifacts are still present, and the details poorly defined. Whereas, SelectionGAN generates better images, by incorporating finer street lines and generating better more realistic trees appearance. PAGAN produces superior images in terms of quality and details. It incorporates more details appearance of trees, and improved street lines, bringing its generated image closer to the ground truth compared to other methods. In Figure \ref{fig:qualitative_op} we provide a comparison on the OP dataset. As shown in Figure \ref{fig:qualitative_op} (a), the Pix2pix, X-Fork methods suffer from a high-level of artifacts and generate noisy images. While X-Seq and SelectionGAN also exhibit noticeable artifacts, they manage to capture part of building on the right side of the image. The GeoUrban method generates a better generated image in terms of image outlines, making it slightly better-performing than other methods in this regard. PAGAN produces the best generated image in terms of outlines and boundaries, showing improved spatial consistency with the ground truth. However, in the central pert of the generated image, it incorrectly generate trees where a building appear in the ground truth. However, in overall, it generates more realistic street-view image in comparison to the other methods.
In Figure \ref{fig:qualitative_op} (b), the Pix2pix, X-Fork, and SelectionGAN methods  generating artifact-heavy and noisy street-view images. In contrast, the X-Seq and GeoUrban methods generate more defined outlines and capture some street lines. PAGAN outperforms other methods by generating geometrically consistent street-view images, despite that some local details are missing in the output, it successfully include finer street lines compared to the other methods. Figures \ref{fig:qualitative_cvact1} show a qualitative comparison on the CVACT dataset. As shown in Figure \ref{fig:qualitative_cvact1}(a), X-Forks generates a street-view image with noticeable artifacts and poorly shaped objects' outlines. While Pix2pix generates better image compared to X-Fork; however it lacks full geometric consistency. SSVG method generates geometrically consistent street-view images with finer details compared with other methods. In Figure \ref{fig:qualitative_cvact1}(b) the Pix2pix and X-Fork methods generate images with an acceptable level of geometric consistency and local details, such as street lines. In addition SSVG is able to generate the sidewalk in the middle, bringing it closer to the ground truth than other methods. 
On the other hand, Figure \ref{fig:qualitative_cvact2} (a) shows that ControlNet adding details that do not exists in the ground truth, making it deviate from the ground truth. In comparison, the CrossViewDiff method generates a geometrically consistent street-view image. Figure \ref{fig:qualitative_cvact2}(b) shows that while CrossViewDiff generates a street-view image with different sidewalk shape on the right side of the image, it remains almost geometrically consistent, compared to the image generated by ControlNet, which represents a different street layout and introduces details not exist in the ground truth. In Figure \ref{fig:qualitative_cvact2} (c) Controlnet generates buildings, cars, and trees in locations where they do not exist in the ground truth, while CrossViewDiff is able to generate geometrically consistent street-view with finer-details.  From this compression, we can conclude that the diffusion-based methods are capable to generate street-view images with  a good level of geometrically consistency and fine-grained details. However, they may sometimes over-detail the output images, as seen with ControlNet. While ControlNet is a general-purpose diffusion model and not discussed in our methods section, we present its qualitative results here for the sake of completeness. It was reported in \cite{li2024crossviewdiff} within the context of satellite-to-street-view generation.}

\section{Quantitative Comparison}
\textcolor{black}{
This section provide a quantitative comparison of the reviewed methods, focusing on their performance across different datasets. Due to the variation in the utilized datasets and  the employed evaluation metrics,  direct comparison between some methods are limited. Therefore, we focused to highlight the commonly reported evaluation metrics for a fair comparison, as shown in Tables \ref{tab:quantitative-comparison-cvusa}, \ref{tab:quantitative-comparison-dayton}, and \ref{tab:quantitative-comparison-cvact}.}
\textcolor{black}{
On the CVUSA dataset, the results show CrossMLP and LGGAN achieved the best highest performance in terms of SSIM, which indicates their strong ability to generate high-quality and structurally well-defined street-view images. SSVS and PanoGAN followed closely with SSIM scores of 0.447 and 0.444, respectively, which reflects a reasonable level of structural similarity with the ground truth. In contrast, SSVG method obtained the lowest  SSIM score with value of 0.3408, suggesting  its poor capabilities of preserving structural elements like image texture and pattern compared to other methods.  For the PSNR metric, CrossMLP reached to the highest score, followed by LGGAN with a score of 22.58. This highlights both methods' low pixel-level reconstruction error. PGSL underperformed other methods in this metric, indicating larger difference from the ground truth. In terms of SD, LGGAN and CrossMLP methods show their high capabilities in generating clear details images with a small performance gap of 0.16. PanoGAN obtained competitive SD score, indicating it maintains comparable level of details and edge sharpness. Finally for KL score, LGGAN demonstrated its high capabilities in generating features that are close to the ground truth distribution. On the other hand, Pix2pix and X-Fork exhibited the highest KL scores, indicating that their generated images deviate from the ground truth distribution. }
\textcolor{black}{
Using the Dayton dataset, SelectionGAN and CrossMLP outperformed other methods in terms of SSIM with values of 0.593 and 0.559, respectively. These values indicate better structural similarity to the ground truth. Both methods also achieved the highest PSNR score with a small gap of 0.26, indicating low pixel-level differences with the ground truth. For SD metric, SelectionGAN achieved to the highest score, demonstrating its ability to generate sharper images with well-defined edges. Regrading the KL divergence, LGGAN outperformed other methods by achieving the lowest score,  indicating its ability to generated images that are closest to the  ground truth distribution and appear more realistic. SelectionGAN and CrossMLP performed well in KL metric, with differences of 0.56 and  0.15 from the best score, respectively. This small difference show that their generating images are relatively close to the ground truth distribution. In contrast, Pix2pix emerged as an outlier with the highest KL score, showing that it generates less realistic images that diverge from the original distribution. 
In the case of the CVACT dataset, SSVG achieved the best SSIM score with value of 0.421, revealing its strong ability to preserve the structural similarity in the generated street-view images. CrossViewDiff  followed closely, highlighting comparable performance in maintaining texture and pattern quality. On the other hand, Pix2pix and X-Fork underperformed in preserving the structural elements compared to the other methods. For PSNR, SSVG and Pix2pix achieved the highest scores with a minimal gap of 0.18, implying that these two methods generate street-view images with low pixel-level errors and high visual quality. For SD, SSVG outperformed others in generating sharpest images with clear details. To ensure consistency across dataset results tables, the KL column is included for CVACT in Table \ref{tab:quantitative-comparison-cvact}. However, KL scores were not reported \cite{shi2022} and \cite{li2024crossviewdiff} from which the results were obtained, so the column is left empty.  }

\textcolor{black}{The quantitative results of CVUSA dataset in this section, obtained from several references: Pix2pix, X-Fork, X-Seq, SelectionGAN, and PAGAN taken from \cite{wu2022cross}. SSVS and PGSL taken from \cite{toker2021}. CrossMLP from \cite{ren2021cascaded}, 
PGSL and SSVS are from \cite{toker2021}. SSVG from \cite{shi2022}, LGGAN from \cite{Tang_2020_CVPR}, PanoGAN from \cite{wu2022_panogan}, and CrossViewDiff from \cite{li2024crossviewdiff}. For Dayton dataset, Pix2pix, X-Fork, X-Seq, SelectionGAN, and CrossMLP taken from \cite{ren2021cascaded}, while LGGAN taken from \cite{Tang_2020_CVPR}. For CVACT dataset, Pix2pix, X-Fork, SSVG results are taken from \cite{shi2022}, and CrossViewDiff taken from \cite{li2024crossviewdiff} }

\begin{table*}[tbp]
\caption{Quantitative comparisons on the CVUSA dataset.}
\label{tab:quantitative-comparison-cvusa}
\begin{center}
\resizebox{15cm}{!}{
\begin{tabular}{lcccclccclcccl}
\hline \hline

\multicolumn{1}{c}{} & SSIM $\uparrow$ & PSNR $\uparrow$ &SD $\uparrow$ & KL $\downarrow$                      &                      & \multicolumn{1}{c}{}                     \\ 
\multicolumn{1}{c}{\multirow{-2}{*}{Method}} &  \\ \hline
Pix2pix \cite{isola2017} &    0.3847 &  18.0329 & 18.1152   & 14.37±1.74 \\ X-Fork \cite{regmi2018} &  0.3698 & 18.6153 & 17.7286 & 14.37±1.74            \\

X-Seq \cite{regmi2018} & 0.3737 & 18.9364 & 17.7544 & 10.72±1.69 \\
SelectionGAN \cite{tang2019multi} & 0.4172 & 19.3237& 18.7789 & 9.18±1.63 \\ 

CrossMLP~\cite{ren2021cascaded}                                  & 0.5251                 & 23.1532                & 19.58                & 2.69 ± 0.94  \\ 
SSVS \cite{toker2021} & 0.447 & 13.895 & 15.221 & 3.59 ± 0.92 \\

PAGAN \cite{wu2022cross} & 0.4315 & 19.4518 & 18.4489 & 7.02±1.62 \\
SSVG \cite{shi2022} & 0.3408 & 13.77 & 16.27 & - \\

PGSL \cite{zhai2017} & 0.414 & 11.502 & 10.631  & - \\

LGGAN \cite{Tang_2020_CVPR} & 0.5238 & 22.5766 & 19.7440 & 2.55 ± 0.95 \\
PanoGAN \cite{wu2022cross} & 0.4437 & 20.9467 & 19.0913 & 4.20±1.19  \\
CrossViewDiff \cite{li2024crossviewdiff} & 0.371 & 12.00 & 16.31 & - 
\end{tabular}}
\end{center}

\end{table*}


\begin{table*}[tbp]
\caption{Quantitative comparisons on the Dayton dataset.}
\label{tab:quantitative-comparison-dayton}
\begin{center}
\resizebox{15cm}{!}{
\begin{tabular}{lcccclccclcccl}
\hline \hline

\multicolumn{1}{c}{} & SSIM $\uparrow$ & PSNR $\uparrow$ &SD $\uparrow$ & KL $\downarrow$                    &                      & \multicolumn{1}{c}{}                     \\ 
\multicolumn{1}{c}{\multirow{-2}{*}{Method}} &  \\ \hline
Pix2pix \cite{isola2017} & 0.4180 & 17.6291 & 19.2821 & 38.26 ± 1.88 \\
X-Fork \cite{regmi2018} & 0.4963 & 19.8928 & 19.4533 & 6.00 ± 1.28 \\
X-Seq \cite{regmi2018} & 0.5031 & 20.2803 & 19.5258 & 5.93 ± 1.32\\
SelectionGAN \cite{tang2019multi} & 0.5938 & 23.8874 & 20.0174 & 2.74 ± 0.86 \\
LGGAN \cite{Tang_2020_CVPR} &0.5457 & 22.9949 & 19.6145 & 2.18 ± 0.74\\ 
CrossMLP \cite{ren2021cascaded} & 0.5599 & 23.6232 & 19.6688 & 2.33 ± 0.80 \\
\end{tabular}}
\end{center}

\end{table*}


\begin{table*}[tbp]
\caption{Quantitative comparisons on the CVACT dataset.}
\label{tab:quantitative-comparison-cvact}
\begin{center}
\resizebox{15cm}{!}{
\begin{tabular}{lcccclccclccc}
\hline \hline

\multicolumn{1}{c}{} & SSIM $\uparrow$ & PSNR $\uparrow$ &SD $\uparrow$ &  KL $\downarrow$ &                     & \multicolumn{1}{c}{}                     \\ 
\multicolumn{1}{c}{\multirow{-2}{*}{Method}} &  \\ \hline
Pix2pix \cite{isola2017} & 0.373 & 14.47 & 16.06 & - \\
X-Fork \cite{regmi2018} & 0.370 & 14.17 & 15.89 & - \\
SSVG \cite{shi2022} & 0.421 & 14.65 &16.33 &-  \\
CrossViewDiff \cite{li2024crossviewdiff} & 0.412 & 12.411 & 16.29 & -\\
\end{tabular}}
\end{center}

\end{table*}

\textcolor{black}{\section{Guidelines for Datasets Creation}}

\textcolor{black}{This section presents guidelines for creating high- quality and diverse satellite-street-view datasets that address the challenges mentioned in Section 8.}

\begin{itemize}

    \item \textcolor{black}{Collect high quality satellite and street-view image pairs, discarding images with poor contrast, blurriness, and distortion.}
    \item \textcolor{black}{Store metadata about the satellite and street-view pairs, including latitude and longitude, providing additional contextual information that support synthesis and geo-localization tasks.}
    \item \textcolor{black}{Collect satellite and street-view image pairs with their corresponding auxiliary  information such as segmentation maps, text prompt, satellite images with LiDAR, and 3D point cloud. These additional input will enhance the street-view generation process and maintaining geometric consistency.}

    \item \textcolor{black}{Collect satellite and street-view image pairs from a wide geographical area, including developed and developing regions. Expanding data collection beyond the main cities helps to ensure data diversity and helps to minimize the data disparities.}
    
    \item \textcolor{black}{Within city, ensure the dataset capture diverse locations, extending the data collection beyond the famous landmarks and mains roads, to capture various street patterns.}

    \item \textcolor{black}{Ensure the collected dataset contains a diverse and balanced distribution of images across environmental settings (rural, urban and sub-urban) to minimize the data biases and improve generalization and robustness of the deep learning models. }
    \item \textcolor{black}{Consider collecting images with time-of-day variation and under different weather conditions (e.g. rain, snow, and spring-time). Additionally, including dynamic scenes like pedestrians vehicles or traffic will help to enrich dataset diversity and ensure model robustness. }

    \item \textcolor{black}{Develop open-access benchmark datasets is imperative for advancing satellite to street-view domain. It facilitates objective performance evaluation of deep learning models, reproducibility and fair comparison between deep learning models. In addition, publicly available datasets attract contributions from wider research community, offering interdisciplinary collaboration, resulting in the enrichment of satellite to street-view research area.}

     \item \textcolor{black}{Create synthetic datasets that simulate extreme conditions such as  occlusion, low-light environments, and noise, to effectively evaluate the robustness of developed models under these challenging real-life scenarios.}

\end{itemize}

\textcolor{black}{In addition to the discussed guidelines, fostering collaborations between academic institutions and organizations, such as Google Maps and Open Street Map, plays an important role in enhancing the datasets diversity and accessibility. These collaborations provide researchers with access large-scale, high-quality, and real-world datasets. Also, it helps to reduce the cost and challenges related to datasets collections.}

\section{Ethical and Privacy Considerations}

\noindent\textcolor{black}{Street-view images have becomes an important source of data in several studies and plays an important role for several applications such as urban planning and autonomous driving. However, as it uses a real-world data, it arises ethical consideration and privacy concerns related to street-view usage and synthesis. In this section we presents some of ethical and privacy concerns related to both real and synthesized street-view images. Street-view images capture real-life information that contains private properties, and potential sensitive locations. The presence of such image in datasets without owners' consent arises ethical concerns about using these images in commercials or research domains. Also, street-view images, whether real or synthesized, can be misused in several contexts such as used for surveillance. It can be integrated with some other data sources, to construct detailed profiles about people, leading to privacy violation and  misuse beyond their intended purposes. Moreover, the synthetic street-view images could be used for misinformation and fake mapping by creating non-existing locations to mislead users --- for example, in real-estate and news reporting. From the privacy perspective, street-view images expose private information include individuals' faces, private residences, license plate, and potential sensitive locations. Even though the real street-view images blur some sensitive information, some synthetics street-view images are generated with high realism, accurately generating realistic images of people or vehicles, leading to privacy violations.Furthermore, once street-view images are combined with other data such as satellite images and home addresses, they can open opportunities for criminals to  plan and identify the weak security points in business and homes. } \noindent \textcolor{black}{To address these ethical and privacy concerns, the following strategies should be considered: }

\begin{itemize}
    \item \textcolor{black}{Advanced Anonymization: ensure an affective and advanced methods are used to obscure and blur sensitive areas and identifiable information such as not limited to, individuals' faces, license plates, and home addresses.}
    \item \textcolor{black}{Data Usage Consent: establishing clear consent protocol is important, to ensure that individuals are aware about how their data is being used.}

    \item \textcolor{black}{Regularity Framework: encourage the development of guidelines and policies that govern the use and collect of street-view images . This include compliance with legal regulations such as GDPR (EU) and CCPA (California), government regulations, and regulation that criminalized the use of fake geographic data for deception. }
    \item \textcolor{black}{Users reporting and Awareness: Enable users to report fake and specious street-view images or any geographic data to prevent fraud. }
\end{itemize}

\section{Real-World Applications}

\textcolor{black}{
Street-view synthesis is useful in several tasks across different domains, including the cross-view geo-localization task, which aims to localize the street-view image by matching it against a collection of geo-referenced satellite images. It plays an important role in numerous applications like augmented reality, autonomous driving, event detection, and robotics \cite{toker2021},\cite{wang2021each}. Since the retrieval component is only able to recognize and locate similar objects, integrating satellite-to-street-view synthesis plays an important role in this task. This helps the generator learn more useful information that enhances the location retrieval accuracy \cite{toker2021}. Also, street-view plays an important role in developing sidewalk GIS data effectively, which is considered a main component in smart city development. To utilize the sidewalk in different smart city applications, researchers require publicly accessible sidewalk GIS data that offers information about the sidewalk’s dimensions, coverage, location, and physical conditions \cite{kang2021developing}, \cite{biljecki2021street}. However, many cities lack such publicly available GIS data for sidewalks. The street-view images have been used for extracting the sidewalk information in the sidewalk detection task, as in \cite{kang2021developing}, while both satellite and street-view images are used in the sidewalk extraction task, like \cite{ning2022sidewalk}. Moreover, street-view images showed their importance in urban analytics and urban planning domains, as they have been used in evaluating urban infrastructure and services \cite{zund2021street} and detecting urban changes \cite{byun2022street}.}

\textcolor{black}{
Deploying satellite to street-view synthesis to resource-contained environments, such as mobile devices and autonomous vehicles, introduces several computational challenges due to limited battery life, computational capabilities, memory, and latency in real-time processing. To addressed these challenges several techniques need to be considered to ensure these developed models can be used effectively in resources-contained environments. The First one is employing light weight architectures, like smaller version of vision transformers (MobileVit), while maintaining high-quality images synthesis. The second one is applying model compression techniques such as quantization, pruning and knowledge distillation. The quantization technique improves the model efficiency by reducing the parameter precision, which minimized the memory usage and speeding up computation. The pruning technique eliminates less important parameters, maintaining model performance while reducing the computational costs. In addition, knowledge distillation technique enables transfer learning from well-trained, large model to a smaller one, allows the smaller model "student" to mimic the performance of the larger one "teacher" \cite{dantas2024comprehensive}. Moreover, hardware optimization techniques can be applied, such as running cloud-based inference for intensive processes, while running the lighter tasks locally. }

\section{Challenges and Limitations}
\label{sec:challenges}
The following sections highlight the challenges and limitations in the Satellite to Street View generation task.

\vspace{1mm}
\noindent
\textbf{Limited available datasets:} The datasets for synthesizing street-view from satellite images are limited in number and not publicly available; they need to be requested from their corresponding authors. This hinders progress; making publicly available datasets will help deliver good-quality research and engage the research community.

\vspace{1mm}
\noindent
\textbf{High computational cost:} Despite the success of recent deep learning methods in training huge models, they require a lot of training data to learn the modality-specific rules or find the relation between the various objects on their own. As a result, it requires a longer training time, a considerable amount of training data, and high computational requirements compared with CNN~\cite{khan2022transformers}. It should be noted that satellite-to-street-view generation requires more data as the relation is very complex due to the entirely different input and output in terms of viewpoint.

\vspace{1mm}
\noindent
\textbf{Evaluation metrics:} Finding appropriate and more specific evaluation metrics for this type of task is challenging. The evaluation metrics applied in the literature, like inception scores, PSNR, SSIM, and SD, are generally for image quality, not specifically for image synthesizing for under-consideration tasks.

\vspace{1mm}
\noindent
\textbf{Lack of multi-modality datasets:} Using a single data modality depending on a single image, like a satellite image, for generating street view is more challenging than using multiple data modalities. However, some of the recent works~\cite{Tang_2020_CVPR,tang2019multi} rely on some auxiliary information like segmentation maps for generating street-view. Regardless, generating segmentation maps may limit the methods if there is any difficulty with its computation, affecting the quality of the synthesized street-view image. Multi-modal datasets might solve the problem and increase the quality of the generation.

\vspace{1mm}
\noindent
\textcolor{black}{\textbf{Datasets Bias:}
 satellite to street-view datasets discussed in this paper shows some biases, including limited geographical coverage and some environment setting are underrepresented. For example, CVUSA dataset focus on rural areas, where urban areas are underrepresented. The CVACT dataset covers wider range of urban and sub-urban areas compared to CVUSA \cite{workman2015wide}, where Dayton dataset \cite{vo2016localizing} cover more urban areas.These biases in the datasets within this domain results in a challenge for training of deep learning models. Once these models trained on dataset biased to specific region, the learned pattern will be limited to those areas, leading to challenges in capturing the minority environment settings.  }

\vspace{1mm}
\noindent
\textbf{Low-resolution layout images:} The satellite images are taken from very far away, only harboring the building’s rooftops with little or no information about small objects like building facades. In addition, many local details, such as vehicles and pedestrians, are not captured in satellite images, which makes it difficult to consider these objects during street view synthesis. Also, in satellite images, places with similar street layouts might appear completely different in the street view, which results in less variation in the synthesized street view images. Moreover, due to the high similarity between road textures and other materials like buildings’ roofs, it is sometimes difficult to distinguish road layouts from buildings’ roofs. To overcome these limitations, images must be taken from a reasonable height containing as much information about the object as possible to help synthesize views true to its street view. 

\vspace{1mm}
\noindent
\textbf{Lack of novel techniques:} The methods used in the literature need to be updated; currently, generative adversarial network and their variants are the most popular choices for a street-view generation. Moreover, various CNNs are employed for feature extraction, while Siamese and Triplet Networks have been used for street-view geo-localization tasks. 

\vspace{1mm}
\noindent
\textbf{Image quality degradations:} The satellite image clarity and quality are usually affected by atmospheric conditions, illumination conditions, and noise. Similarly, the satellite images may also be obscured due to dust and clouds, significantly affecting the quality of the generated street-view images. In some cases, building and tree shadows affect the satellite image clarity, which causes the street layouts to be difficult to recognize. One way to solve the mentioned problem is by adopting shadow removal techniques; although these might be useful, but they will have computational costs.

\vspace{1mm}
\noindent
\textbf{Diverse weather conditions:} In usual image generation techniques, the conditions are generally constant and do not change over the course. However, for this task, the same image may have to be generated under diverse conditions, such as the exact layout being different for day and night, snow and spring seasons and many others. This situation needs to have representative images in the datasets to be created.

One of the key challenges for generative models is maintaining robustness against noisy inputs, which can significantly degrade the quality of extracted features. In many real-world domains, input data are often corrupted due to environmental factors. This issue is particularly prominent in satellite imagery, where noise can result from factors such as cloud cover, sensor defects, and adverse weather conditions. Such noise can have a substantial negative impact on performance in tasks like image synthesis. Therefore, enhancing model robustness to noisy inputs is critical for ensuring the reliability and quality of the generated outputs, as well as improving generalization across diverse real-world scenarios.

Several strategies can be employed to address the issue of noisy inputs. One approach involves applying deep learning-based pre-processing techniques to denoise inputs before they are fed into generative models. For example, Denoising Autoencoders (DAEs) \cite{vincent2008extracting} are trained to reconstruct clean images from noisy versions. Additionally, incorporating auxiliary information—such as geo-spatial metadata or semantic maps—can provide valuable complementary signals when satellite imagery is partially obscured. For instance, when cloud cover limits visual clarity, auxiliary data can help infer the underlying scene. Furthermore, data augmentation that includes noisy samples during training can improve model robustness and generalization, effectively preparing the model to handle noisy data encountered in real-world applications.
  
\section{Future Research Directions}
\textcolor{black}{
 \newline
Despite of progress made in satellite to street-view synthesis domain, there are still some challenges that need to be addressed. To further advance this field, these challenges need to be addressed and future research should focus on the following: }

\begin{itemize}
    \item \textcolor{black}{Researchers must consider and focus on recent state-of-the-art techniques such as transformers and stable diffusion for satellite to street-view synthesizing.
    \item Integrate multi-modal data such as depth maps, text description, and LiDAR to provide richer spatial and contextual information, resulting in improving model accuracy and enhances its generalization capabilities.
    \item In addressing the lack of multi-modalities, integrating Multi-interactive Feature Learning mechanisms could improve the model's ability to leverage complementary information across data sources (e.g., satellite imagery, semantic maps, and LiDAR). Moreover, the creation of a Full-time Multi-modality Benchmark—one that includes synchronized multi-modal data under diverse real-world conditions—would provide a standardized foundation for evaluating image fusion and segmentation methods.
    \item For the challenge of low-resolution layout images, incorporating super-resolution techniques presents a promising avenue. These methods can be applied either as pre-processing steps or embedded within generative frameworks to enhance spatial detail, which is crucial for downstream tasks such as fine-grained localization and image synthesis. Exploring cross-domain super-resolution approaches, particularly those adapted to the unique characteristics of satellite and layout imagery, may significantly improve model performance in this context.
    \item Create high-quality, large-scale, divers, and open-access datasets, as recommended Section 5 (Guidelines of Datasets Creation), to enrich research in this area. 
    \item Employ cross-dataset training, to expose  model to a divers environmental settings, which improves the model's generalization.
    \item Transformers and diffusion model have shown promising results in satellite to street-view task. In the future, leveraging these models or their variants more extensively in this domain offer valuable opportunities. Also, integrating Large Language model (LLMs) with the diffusion models could further enhance the generation process and helps to get enrich information and generate a descriptive prompt for satellite images to guide the diffusion model. This integration can help to provide contextual understanding and leading to generating more realistic and accurate street-view images. The diffusion models are capable of generating street-view images with high-quality and fine-grained details. However, they sometimes generate  a bit generic images and may lack full geometric consistency with the ground truth. Adopting a hybrid approach that combines diffusion model and GAN could help to improve the overall framework by leveraging the strengths of both models. This incorporation enables to learn more robust features, resulting in improved street-view generation.}

\end{itemize}
\section{Conclusion}
\textcolor{black}{
This paper has surveyed the satellite to street-view synthesis domain from different aspects, including the recently available methods, common datasets (benchmark datasets), common evaluation metrics, challenges and limitations, by examining 23 recently published papers.
It is clear from the reviewed papers that the Generative Neural Network (GAN) and its varieties are very immersed and widely used in this domain, despite other novel methods being available, diffusion and transformer models have received little attention in this domain.  Despite the success of the reviewed methods, they still struggle to generate small and detailed objects,  which confirms that novel methods need to be used to overcome these limitations and generate more realistic and detailed street-view images. In addition, the research highlights that insufficient diverse publicly available datasets, and most of the surveyed papers rely on limited datasets. This indicates the need for more high-quality, publicly available datasets to enrich the research in this domain. Moreover,  methods in the literature have been evaluated on image quality-based metrics, indicating the importance of more specific evaluation metrics to evaluate the generated images. Overall, the disclosed research challenges highlight that the satellite to street-view synthesis is still an active area of research. Considering the interdisciplinary nature of this field, effectively addressing these challenges will require cross-disciplinary collaborations between geospatial science, urban planning, and machine learning to foster the development of advanced methods and utilize diverse datasets. The developments in satellite to street-view synthesis domain offer considerable policy implications in several domains, including environmental management, urban planning, disaster response and smart cities. By leveraging up-to-date and high-resolution data, policymakers can gain valuable insights and improve the decision-making processes in areas like infrastructure evaluation and climate change adaptation. It is essential for academic institutions, governments, and industries to collaborate in advancing and deploying these models effectively in real-world applications. We suggest to integrating this domain into education curricula to prepare future researchers and train them in geospatial AI, for advancing this domain. This integration will equip researchers with the required skills to address challenges, ultimately enhancing applications in several domains such as urban planning and autonomous driving. We also encourage researchers to contribute in addressing the outlined challenges and to foster research collaborations to drive progress and innovations in this research area.  }


\bigskip




\vspace{2mm}
\noindent
\textbf{Acknowledgement:}
\textcolor{black}{A preprint of this paper is available on arXiv at \url{https://arxiv.org/abs/2405.08961}.} The author/s would like to acknowledge the support received from Saudi Data and AI Authority (SDAIA) and King Fahd University of Petroleum and Minerals (KFUPM) under SDAIA-KFUPM Joint Research Center for Artificial Intelligence Grant No. JRC-AI-RFP-11.



\bibliography{sn-bibliography} 

\end{document}